\PassOptionsToPackage{unicode}{hyperref}
\PassOptionsToPackage{hyphens}{url}
\PassOptionsToPackage{dvipsnames,svgnames,x11names}{xcolor}
\documentclass[
]{article}

\usepackage{tcolorbox}
\usepackage{amsmath,amssymb}
\usepackage{iftex}
\ifPDFTeX
  \usepackage[T1]{fontenc}
  \usepackage[utf8]{inputenc}
  \usepackage{textcomp} 
\else 
  \usepackage{unicode-math}
  \defaultfontfeatures{Scale=MatchLowercase}
  \defaultfontfeatures[\rmfamily]{Ligatures=TeX,Scale=1}
\fi
\usepackage{lmodern}
\ifPDFTeX\else  
\fi
\IfFileExists{upquote.sty}{\usepackage{upquote}}{}
\IfFileExists{microtype.sty}{
  \usepackage[]{microtype}
  \UseMicrotypeSet[protrusion]{basicmath} 
}{}
\usepackage{xcolor}
\usepackage[left=1.25in,right=1.25in]{geometry}
\makeatletter
\ifx\paragraph\undefined\else
  \let\oldparagraph\paragraph
  \renewcommand{\paragraph}{
    \@ifstar
      \xxxParagraphStar
      \xxxParagraphNoStar
  }
  \newcommand{\xxxParagraphStar}[1]{\oldparagraph*{#1}\mbox{}}
  \newcommand{\xxxParagraphNoStar}[1]{\oldparagraph{#1}\mbox{}}
\fi
\ifx\subparagraph\undefined\else
  \let\oldsubparagraph\subparagraph
  \renewcommand{\subparagraph}{
    \@ifstar
      \xxxSubParagraphStar
      \xxxSubParagraphNoStar
  }
  \newcommand{\xxxSubParagraphStar}[1]{\oldsubparagraph*{#1}\mbox{}}
  \newcommand{\xxxSubParagraphNoStar}[1]{\oldsubparagraph{#1}\mbox{}}
\fi
\makeatother

\usepackage{longtable,booktabs,array}
\usepackage{calc} 
\usepackage{etoolbox}
\makeatletter
\patchcmd\longtable{\par}{\if@noskipsec\mbox{}\fi\par}{}{}
\makeatother
\IfFileExists{footnotehyper.sty}{\usepackage{footnotehyper}}{\usepackage{footnote}}
\makesavenoteenv{longtable}
\usepackage{graphicx}
\makeatletter
\def\maxwidth{\ifdim\Gin@nat@width>\linewidth\linewidth\else\Gin@nat@width\fi}
\def\maxheight{\ifdim\Gin@nat@height>\textheight\textheight\else\Gin@nat@height\fi}
\makeatother
\setkeys{Gin}{width=\maxwidth,height=\maxheight,keepaspectratio}
\makeatletter
\def\fps@figure{htbp}
\makeatother

\usepackage{orcidlink}
\definecolor{mypink}{RGB}{219, 48, 122}
\makeatletter
\@ifpackageloaded{caption}{}{\usepackage{caption}}
\AtBeginDocument{%
\ifdefined\contentsname
  \renewcommand*\contentsname{Table of contents}
\else
  \newcommand\contentsname{Table of contents}
\fi
\ifdefined\listfigurename
  \renewcommand*\listfigurename{List of Figures}
\else
  \newcommand\listfigurename{List of Figures}
\fi
\ifdefined\listtablename
  \renewcommand*\listtablename{List of Tables}
\else
  \newcommand\listtablename{List of Tables}
\fi
\ifdefined\figurename
  \renewcommand*\figurename{Figure}
\else
  \newcommand\figurename{Figure}
\fi
\ifdefined\tablename
  \renewcommand*\tablename{Table}
\else
  \newcommand\tablename{Table}
\fi
}
\@ifpackageloaded{float}{}{\usepackage{float}}
\floatstyle{ruled}
\@ifundefined{c@chapter}{\newfloat{codelisting}{h}{lop}}{\newfloat{codelisting}{h}{lop}[chapter]}
\floatname{codelisting}{Listing}

\makeatother
\makeatletter
\@ifpackageloaded{caption}{}{\usepackage{caption}}
\@ifpackageloaded{subcaption}{}{\usepackage{subcaption}}
\makeatother

\ifLuaTeX
  \usepackage{selnolig}  
\fi
\usepackage[]{natbib}
\setcitestyle{round}
\usepackage{bookmark}

\IfFileExists{xurl.sty}{\usepackage{xurl}}{} 
\hypersetup{
  pdftitle={Language Models as Models of Language},
  pdfauthor={Raphaël Millière},
  pdfkeywords={Language models, Linguistics, Syntax, Nativism},
  colorlinks=true,
  linkcolor={blue},
  filecolor={Maroon},
  citecolor={Blue},
  urlcolor={Blue},
  pdfcreator={LaTeX via pandoc}}

\title{Language Models as Models of Language}
\author{
Raphaël Millière\\Macquarie
University\\\href{mailto:raphael.milliere@mq.edu.au}{raphael.milliere@mq.edu.au}}
\date{}
\begin{document}
\maketitle

\noindent
\begin{tcolorbox}[
  colback=white,
  colframe=black,
  boxrule=0.5pt,
  left=6pt,
  right=6pt,
  top=6pt,
  bottom=6pt,
  arc=0pt,
  before skip=12pt,
  after skip=12pt
]
\centering
Forthcoming in Nefdt, R., Dupre, G., \& Stanton, K. (eds.), \textit{The Oxford Handbook of the Philosophy of Linguistics}. Oxford University Press.
\end{tcolorbox}

\vspace{1em}

\begin{abstract}
\noindent This chapter critically examines the potential contributions of modern
language models to theoretical linguistics. Despite their focus on
engineering goals, these models' ability to acquire sophisticated
linguistic knowledge from mere exposure to data warrants a careful
reassessment of their relevance to linguistic theory. I review a growing
body of empirical evidence suggesting that language models can learn
hierarchical syntactic structure and exhibit sensitivity to various
linguistic phenomena, even when trained on developmentally plausible
amounts of data. While the competence/performance distinction has been
invoked to dismiss the relevance of such models to linguistic theory, I
argue that this assessment may be premature. By carefully controlling
learning conditions and making use of causal intervention methods,
experiments with language models can potentially constrain hypotheses
about language acquisition and competence. I conclude that closer
collaboration between theoretical linguists and computational
researchers could yield valuable insights, particularly in advancing
debates about linguistic nativism.
\end{abstract}

\renewcommand*\contentsname{Table of contents}
{
\hypersetup{linkcolor=}
\setcounter{tocdepth}{3}
\tableofcontents
}

\section{Introduction}\label{sec-intro}

The recent success of artificial neural networks in natural language
processing has sparked renewed interest in their potential to elucidate
longstanding questions in linguistics. Modern neural networks based on
deep learning architectures and trained on linguistic data, called
\emph{language models}, now match or exceed human performance on many
language tasks once thought intractable for machines. Historically,
connectionist models were critiqued as merely statistical approximations
of linguistic behaviour, fundamentally unable to capture the underlying
competence of human language users. While the remarkable progress of
modern language models has been largely driven by engineering efforts
rather than research goals, it nonetheless warrants a careful
re-examination of the relevance of neural networks for linguistics as a
scientific field.

This chapter aims to critically examine what language models may
contribute -- if anything -- to theoretical linguistics. In particular,
it considers whether we should take language models seriously \emph{as
models of language}; or, more precisely, as models of human language
acquisition and competence. Section~\ref{sec-brief-history} provides a
brief historical overview of statistical language modelling in natural
language processing research, from early experiments inspired by
information theory to modern language models based on the Transformer
architecture. Section~\ref{sec-syntax} turns to the rich body of work in
computational linguistics investigating the linguistic knowledge of
modern language models, with a particular focus on syntax. This line of
research involves probing the sensitivity of language models to
syntactic features through linguistically-informed experiments. The
implications of these empirical findings for theoretical linguistics are
discussed in Section~\ref{sec-language-models-linguistics}. This section
examines three potential interpretations of language models: as models
of linguistic performance, competence, and acquisition. It is often
assumed language models merely capture patterns of usage rather than the
abstract linguistic competence underlying language. However, insights
from computational linguistics increasingly suggest that language models
trained in plausible learning scenarios and tested in carefully
controlled conditions may serve as fruitful testbeds for evaluating
linguistic hypotheses. In particular, language models show promise as
idealized model learners to test or constrain theories of language
acquisition and make headway on ongoing debates about linguistic
nativism. Available evidence remains tentative, however; fully realizing
the potential of language models to inform linguistic theory will likely
require embracing open-minded collaboration between computational and
theoretical linguists.\footnote{This chapter is designed to accommodate
  readers with varying levels of background knowledge and interests.
  Readers already familiar with the history of statistical language
  modeling may wish to skip Section~\ref{sec-brief-history}. Readers who
  are already well-acquainted in recent empirical work on language
  models' linguistic abilities may also wish to skip
  Section~\ref{sec-syntax} and proceed directly to
  Section~\ref{sec-language-models-linguistics}, which explores the
  theoretical and philosophical implications of this research for
  linguistics. Those primarily interested in the debate surrounding
  language models as scientific models of language may find that section
  most relevant.}

\section{A brief history of statistical language
modelling}\label{sec-brief-history}

\subsection{Early efforts}\label{early-efforts}

Natural language processing (NLP) traces its origins to the late 1940s
and early 1950s, with early attempts to develop computer programs
capable of processing and understanding human language. This research
programme was inspired by the advent of the first programmable digital
computers, raising hopes that complex linguistic tasks, such as language
translation, could potentially be replicated algorithmically. From the
beginning, the history of NLP was marked by a tension between two
competing approaches: a stochastic approach influenced by information
theory, and a symbolic approach influenced by theoretical linguistics.

In 1948, Claude Shannon proposed a probabilistic model of communication,
expressing the information content of a message in terms of its
probability \citep{shannonMathematicalTheoryCommunication1948}. Although
Shannon's work focused primarily on telecommunication, his methods of
measuring information entropy found applicability in understanding
linguistic phenomena. In fact, the first application of statistical
methods to NLP is credited to Shannon himself, who experimented with
different techniques for predicting the next letter in a sequence of
English text based on the preceding letters
\citep{shannonPredictionEntropyPrinted1951}. Shannon's theory also
inspired other researchers to tackle challenging problems in NLP through
statistical methods. In a memorandum published in 1949, for example,
Warren Weaver proposed to use information theory as a framework for
machine translation \citep{weaverTranslation1955}. By determining the
statistical regularities between two languages, Weaver postulated that
one could find an optimal mapping between them to enable translation.

Despite these early efforts, NLP research was initially dominated by
symbolic rather than statistical methods, as linguistic theory inspired
efforts to explicitly encode linguistic rules for machines. Noam
Chomsky's work was particularly influential on this development
\citep{chomskySyntacticStructures1957, chomskyAspectsTheorySyntax1965}.
Rather than seeing language as a set of learned habits or responses to
stimuli, Chomsky argued that our ability to generate and understand an
infinite number of sentences suggests that language use must be involve
the algorithmic manipulation of hierarchical symbolic structures
according to unconscious grammatical rules. On his view, these
unconscious rules were acquired by language learners through reliance on
a posited innate language faculty, dubbed ``universal grammar'', which
constrained the space of possible human languages. The idea that
linguistic knowledge could be viewed as an abstract deductive system
inspired precise formalisms that could be translated into symbolic rules
for computers.

Many NLP projects adopted this linguistics-driven approach in the 1960s
and 1970s, hand-engineering complex symbolic rule systems to parse input
and generate responses using limited vocabularies. Terry Winograd's
SHRDLU, for instance, used a form of syntactic parsing to break down
English sentences into subject-verb-object chunks and translate them
into action commands in a simplified ``blocks world''
\citep{winogradProceduresRepresentationData1971}. Other systems
explicitly incorporated insights from Chomskyan linguistics. LUNAR, for
example, was designed to be a natural language interface that could
answer questions about Apollo 11 moon rock samples for NASA
\citep{woodsProgressNaturalLanguage1973}. Inputs were processed using an
``augmented transition network'' inspired by Chomsky's transformational
grammar, that could recursively apply transformation rules to parse
English questions into a deep structural representation.

While the symbolic approach was initially fruitful, it also showed
significant limitations that proved difficult to overcome. Rule-based
NLP algorithms were labour-intensive to create, often brittle in the
face of linguistic variability, and struggled with the ambiguity
inherent in natural language. Although they could represent complex
linguistic structures, their reliance on hand-crafted rules made them
less flexible and adaptable to different languages and domains, and they
often failed to adequately model the complexities of semantic and
pragmatic context that are integral to human language understanding.

Another influential idea emerged in parallel from structural
linguistics, which aimed to uncover the rules and patterns that govern
language as a system of interrelated symbols. In contrast with Chomsky's
generative linguistics, structuralism focused on describing language as
it is used rather than specifying abstract rules of grammar. In this
context, Zellig Harris suggested that words appearing in similar
contexts are likely to have related or overlapping meaning
\citep{harrisDistributionalStructure1954}. This claim, which came to be
known as the \emph{distributional hypothesis}, was aptly summarized by
J. R. Firth with the slogan ``You shall know a word by the company it
keeps'' \citep{firthSynopsisLinguisticTheory1957}. Firth explicitly
acknowledged the influence of Wittgenstein's conception of meaning as
use, according to which the meaning of a word is derived from its use in
language, rather than from the object it refers to or the mental
representation it is associated with. Building on this intuition, the
distributional hypothesis states that the contextual meaning of words
emerges from their place in linguistic environments and habitual
associations with other words. Firth also emphasized the importance of
analysing large samples of authentic language use to understand meaning,
idiom, and lexicology -- anticipating later developments in statistical
approaches to NLP.

These ideas reached fuller fruition through the development of
quantitative methods to model relationships between words. The work of
Charles Osgood in psychology was a notable precursor in this area
\citep{osgoodNatureMeasurementMeaning1952}. Osgood hypothesized that by
quantifying allocation of concepts along a standardized set of
dimensions, one could measure their meaning. His ``semantic
differential'' method involved presenting subjects with a concept (e.g.,
``dictator'') and a scale between two opposites (e.g., ``kind-cruel'').
The subject would then rate where the concept falls on the scale.
Repeating this for many concepts on many scales located the concepts in
a semantic space. A factor analysis subsequently identified three main
dimensions accounting most of the variance in ratings: \emph{evaluation}
(e.g.~good-bad), \emph{potency} (e.g.~strong-weak), and \emph{activity}
(e.g.~active-passive).

Osgood's ``semantic differential'' experiments introduced the important
idea that meaning could be represented geometrically in a
multidimensional vector space, although it relied on explicit
participant ratings rather than distributional properties of words in a
linguistic corpus. However, subsequent research combined vector-based
representations with a data-driven approach. Much of this work was done
to improve information retrieval: by representing documents as vectors
whose dimensions correspond to words weighted by frequency, one could
search and retrieve documents from the similarities between their
vectors \citep{saltonVectorSpaceModel1975}. The idea of representing
documents or words as vectors in high-dimensional vector spaces, and
using distance between vectors as proxy for semantic similarity, would
become a key principle of statistical approaches to language modelling
building on the distributional hypothesis.

Latent semantic analysis (LSA) emerged in the late 1980s as a more
sophisticated method to uncover latent semantic structure from text
corpora \citep{deerwesterIndexingLatentSemantic1990}. LSA represents
documents as vectors of weighted word frequencies, then applies a
technique called singular value decomposition to \emph{reduce} the
dimensionality of the resulting vector space, while preserving important
semantic information. Importantly, the relationships between words
captured by the similarity between their corresponding vectors in LSA
reflect deeper semantic similarity rather than just surface
co-occurrence statistics. Indeed, the similarity between word vectors
should not be confused with the frequency or likelihood of words
appearing together in the corpus. Rather, the distance between vectors
captures the similarity in the effects the words have on the meaning of
the passages in which they occur. Words that do not directly co-occur in
the text can still have highly similar vectors if they affect passage
meanings in similar ways. This allows LSA to detect semantic
relationships between words without relying solely on co-occurrence
counts -- including synonymy, antonymy, hypernymy, and meronymy.

Interestingly, the success of LSA was not merely seen as an engineering
achievement, but thought to have implications for our understanding of
human cognition. Thomas Landauer and Susan Dumais, two of the creators
of LSA, took it to challenge nativist theories of language acquisition,
arguing that it provides an existence proof that a general statistical
learning mechanism over large corpora can rapidly acquire semantic
knowledge on the scale of children's vocabulary growth
\citep{landauerSolutionPlatoProblem1997}. Specifically, they argued the
high-dimensional vector representations learned by LSA provide a
potential computational mechanism to explain how learners can acquire so
much knowledge from limited linguistic input, by allowing small
adjustments to the representation of each word during new exposures to
propagate across the lexicon, explaining rapid accumulation of
vocabulary. As we shall see, similar claims about language acquisition
have been bolstered by the success of modern language models.

\subsection{Word embeddings models}\label{word-embeddings-models}

Research on distributional semantics reached maturity with the
development of word vector models based on artificial neural networks
\citep{bengioNeuralProbabilisticLanguage2000, mikolovEfficientEstimationWord2013}.
The key insight behind such models is that the distributional properties
of words can be learned by training a neural network to \emph{predict} a
word's context given the word itself, or vice versa. Unlike previous
statistical methods, these neural models encode words into dense,
low-dimensional vector representations also known as \emph{word
embeddings}. The resulting vector space drastically reduces the
dimensionality of linguistic data while preserving information about
meaningful linguistic relationships better than LSA. Word embedding
models demonstrate the ability of statistical methods inspired by the
distributional hypothesis to learn rich representations of lexical
knowledge from unlabelled text. As the neural network trains on large
amounts of text, words with similar meanings and syntactic roles
converge to similar embedding locations that support predicting their
shared contexts.

A particularly influential technique known as Word2Vec demonstrated the
power of word embedding models to capture both semantic and syntactic
regularities from their training data
\citep{mikolovEfficientEstimationWord2013}. One of the key insights from
Word2Vec is that analogical relationships between words could be
captured by simple arithmetic operations on their vector representations
in the vector space of the trained model. For example, subtracting the
vector for ``man'' from the vector for ``king'' then adding the vector
for ``woman'' would result in a vector closest to the vector for
``queen'' in the space -- implicitly capturing the idea that ``man'' is
to ``king'' what ``woman'' is to ``queen''. The vector space of Word2Vec
models also exhibits morphological relationships between word forms. For
example, the vector offsets between ``walk'' and ``walked'' versus
``swim'' and ``swam'' are parallel in the vector space, suggesting it
can capture regular rules of inflectional morphology (e.g.,``walk'' is
to ``walked'' what ``swim'' is to ``swam''). Relationships between
derivationally related words can also be captured. For example, even
though morphemes like ``-er'' don't occur as standalone units in text,
Word2Vec models appear to represent them implicitly, mirroring
derivational morphology: vector offsets akin to ``walk'' - ``walker'' +
``swim'' result in a vector closest to ``swimmer''.

This ability of word embeddings models to represent nuanced lexical
relationships points to their potential to inform linguistic theory
\citep{lenciDistributionalModelsWord2018}. Classical approaches to
lexical semantics often treat word meaning as a combination of binary
semantic features; for example, the word ``bachelor'' carries the
features {[}+human{]}, {[}+male{]}, {[}+unmarried{]}
\citep{katzStructureSemanticTheory1963}. But modelling word meaning in
this way seems intractable in practice, as the lexicon of any natural
language contains a dizzyingly large number of distinct word meanings
\citep{baroniFregeSpaceProgram2014}. Word embedding models can
automatically acquire meaning representations from corpus data, scaling
up to large lexicons in a way not feasible for manually created
representations. They provide an empirically motivated way to model
meaning that captures gradience and flexibility; for example, they
naturally represent vagueness through graded similarities
\citep{erkProbabilisticTurnSemantics2022}.

Word embedding models can be used in exploratory ways to uncover
patterns in large-scale distributional data like word similarities and
nearest neighbours \citep{boledaDistributionalSemanticsLinguistic2020}.
Specific linguistic phenomena can be investigated by looking at
distributional representations. For example, word embeddings in models
trained on linguistic data from different time periods can provide
useful insights about semantic change over time
\citep{kimTemporalAnalysisLanguage2014, hamiltonDiachronicWordEmbeddings2016}.
Furthermore, word embedding models can be used to evaluate linguistic
hypotheses by translating them into distributional terms and testing
their predictions \citep{boledaDistributionalSemanticsLinguistic2020}.
For example, \citet{boledaIntensionalityWasOnly2013} tested the
hypothesis that adjectives expressing modality (e.g., ``alleged'' or
``possible'') are harder to model compositionally than non-modal
adjectives, by looking at distances between corresponding vectors.
Instead of confirming their hypothesis, they found that typicality of
the adjective-noun pairing was more predictive of compositionality,
leading them to propose that composition relies jointly on conceptual
typicality and referential context
\citep{mcnallyConceptualReferentialAffordance2017}.

This line of research inspired ongoing efforts to combine insights from
distributional semantics with formal semantics
\citep{erkSemanticsDistributionalRepresentations2013, boledaFormalDistributionalSemantics2016, venhuizenDistributionalFormalSemantics2022}.
Formal semantics excels at modelling phenomena like quantification,
negation, modality, and logical inference but struggles with lexical
semantics and descriptive content. Distributional semantics has
complementary strengths -- it captures lexical and conceptual meaning
very well through the statistical analysis of large corpora, but cannot
easily handle function words or logical entailments. This creates an
incentive to combine both approaches into an integrative semantic
framework called ``Formal Distributional Semantics''. However, this is
challenging due to the fundamentally different theoretical foundations
of each approach. One strategy consists in enhancing formal semantics
with distributional information that acts as a supplementary layer over
logical form \citep{beltagyMontagueMeetsMarkov2013}. An alternative
strategy starts instead from distributional semantics, and aims to
reformulate logical phenomena like quantification directly in terms of
operations over distributional vectors, without relying on an existing
logic \citep{herbelotBuildingSharedWorld2015}. Each strategy faces
distinct challenges: the former struggles to integrate distributional
lexical knowledge into existing formal logics in a coherent way,
reconcile vector similarities with formal inference, and retain
cognitive plausibility; the latter has difficulty recovering logical
phenomena like quantification directly from distributional spaces and
lacks a clear notion of reference. Nonetheless, Formal Distributional
Semantics is a promising avenue of research that vividly illustrates the
relevance of distributional models to theoretical linguistics.

\subsection{Language models}\label{language-models}

Despite their success in modelling salient aspects of lexical
relationships, word embedding models have several significant
limitations. Firstly, they assign a single ``static'' vector
representation to each word type, which prevents them from modelling
variations in word meaning based on context or disambiguating homonyms.
Secondly, they rely on ``shallow'' neural network architectures
(typically with a single hidden layer), which may limit their ability to
capture complex hierarchical relationships between words. Finally, these
models fundamentally treat language as a mere ``bag of words,''
disregarding information about word order. Being designed to model
language at the level of individual words, they are ill-suited to
represent complex linguistic expressions. While it is possible to
compute a vector representation for a complex expression by averaging
the vectors of the words it contains, this fails to capture information
about compositional structure.

These shortcomings are addressed by modern language models based on deep
neural networks.\footnote{See
  \citet{millierePhilosophicalIntroductionLanguage2024} for an
  introduction to neural language models.} By contrast with shallow
models, these neural networks have many hidden layers, affording them
greater representational flexibility
\citep{lecunDeepLearning2015, bucknerDeepLearningPhilosophical2019}.
Unlike word embedding models, they model the meaning of individual words
in context, and can process complex whole sentences or paragraphs while
preserving information about word order and syntactic structure. In
recent years, these models have taken over virtually every corner of
NLP, demonstrating unprecedented performance on a wide array of
linguistic tasks that were previously challenging even for task-specific
models
\citep{brownLanguageModelsAre2020, openaiGPT4TechnicalReport2023}.

Virtually all modern language models use a deep neural network
architecture called the Transformer \citep{vaswaniAttentionAllYou2017}.
The most common variant of the Transformer (known as ``decoder-only'' or
``autoregressive'' Transformer) learns from \emph{next-word prediction}:
given a sequence of words \(w_1, w_2,..., w_i\) passed as input to the
model, it attempts to predict the subsequent word
\(w_{i+1}\).\footnote{This description is slightly simplified for the
  sake of exposition. To be precise, autoregressive Transformer models
  process ``tokens'' rather than words. The input sequence is divided
  into these tokens through a process called \emph{tokenization}, aimed
  at maintaining a balance between computational efficiency and the
  capacity to represent diverse words, including complex and less common
  ones. While many tokens do map onto whole words, others map onto
  sub-word units that may or may not carve words at their
  morphologically meaningful joints. For instance, in GPT-3, the word
  ``linguistics'' is tokenized into three separate units: ``ling'',
  ``u'', and ``istics''.} These models are trained on a large amount of
data, by sampling fixed-length sequences for next-word prediction over
and over again. When training begins, the model is no better than chance
at predicting the next word. Each time a prediction is made, however, it
is compared to the word that \emph{actually} follows the input sequence
in the training data. The difference -- or ``error'' -- between the
model's prediction and the ground truth is then used to adjust the
model's internal parameters. These adjustments are calculated such that
they would decrease the prediction error if the same context were
encountered again. Through exposure to vast amounts of text, the model
incrementally refines its performance, learning to predict the next word
in any context occurring in the training data.

The fundamental building block of the Transformer architecture is a
remarkably versatile mechanism known as \emph{self-attention}. In
essence, self-attention allows the model to weigh the relative
importance of different words in the input sequence when predicting a
new word. When a word is processed through the self-attention mechanism,
the model ``attends'' to all the preceding words, gathering relevant
information that might be spread out across the sequence. The mechanism
assigns a weight (called ``attention score'') to each of these words,
determining how much each should contribute to the current prediction.
For instance, when predicting the verb in a sentence, the model might
give a high attention score to the verb's subject, even if it is far
back in the sequence. Conversely, less relevant tokens -- such as a
distant conjunction or adverb -- might receive lower attention scores.
Importantly, the allocation of these scores is not static; it is
adjusted dynamically based on the current prediction task. In practical
terms, these attention scores are used to create a weighted combination
of the word vectors in the sequence, and this combination is then used
to predict the next word. This mechanism ensures that more important
words have a larger influence on the prediction, while less important
words have a lesser influence.

Each self-attention module in a Transformer-based language model is
called an ``attention head''. Language models do not contain a single
attention head, but may have thousands of them.\footnote{GPT-3, for
  example, has 9,216 attention heads: 96 heads in each of 96 layers
  \citep{brownLanguageModelsAre2020}. State-of-the-art models like GPT-4
  plausibly have more, although this information is not publicly
  available \citep{openaiGPT4TechnicalReport2023}.} Each of these
attention heads has trainable parameters, which means that they may
specialize during training to attend to specific kinds of dependencies
between words. This mechanism allows language models to avoid some
issues that plagued previous techniques, like the inability to deal with
long-range dependencies or the difficulty in capturing ambiguity or
context-sensitive influences on meaning. By directly modelling
relationships between all words in a sequence, regardless of their
distance from each other, Transformer-based language models are, in
principle, particularly well-suited to induce syntactic structure.
Importantly, while attention heads themselves process all words in
parallel (i.e., they are ``permutation equivariant''), information about
word order is preserved through a mechanism known as positional
encoding.

Transformer-based language models have shown impressive capabilities
with a very broad range of tasks. They can generate fluent and
grammatically well-formed text in natural language on virtually every
topic. Beyond free-form text generation, they achieve good performance
at summarization, paraphrasing, translation, information retrieval,
sentiment analysis, and question answering, among other classic NLP
tasks. Importantly, they can do so just from being pre-trained on a vast
corpus of text with a next-word prediction learning objective, without
task-specific fine-tuning \citep{brownLanguageModelsAre2020}.

These results have inspired several research programmes in computer
science, computational linguistics and adjacent fields. A first set of
issues relates to the systematic assessment of the capacities and
limitations of language models. In the linguistic domain, in particular,
there are ongoing efforts to investigate what kind of linguistic
knowledge and competence, if any, can be meaningfully ascribed to
language models. A distinct but related set of issues concerns the
potential implications that experiments with language models may have
for theoretical linguistics and developmental psycholinguistics. One
particularly controversial issue is whether the apparent success of
language models in learning the syntax of natural languages without
built-in syntactic knowledge may challenge or constrain theories of
language acquisition.

In what follows, I will consider each set of issues in turn. While
discussing the putative linguistic competence of language models raises
fascinating questions about semantic competence \citep[see
e.g.][]{benderClimbingNLUMeaning2020, sogaardUnderstandingModelsUnderstanding2022, piantadosiMeaningReferenceLarge2022, molloVectorGroundingProblem2023},
I will focus more closely on issues related to syntactic knowledge that
have more straightforward implications for linguistics.

\section{What do language models know about syntax?}\label{sec-syntax}

Large language models like GPT-3 and GPT-4 hardly ever make grammatical
mistakes. In fact, these models can reliably generate whole paragraphs
of syntactically coherent text adapted to the style, tone, and language
of the input. On the face of it, this seems to imply that they have
effectively learned the underlying rules and structure of natural
language syntax from the linguistic data they were trained on. This
could be taken as preliminary evidence that, given sufficient parameters
and training examples, neural networks can acquire sophisticated
knowledge about core aspects of human language like hierarchical phrase
structure and compositionality purely through exposure, without explicit
supervision. However, the mere fact that language models can generate
grammatical sentences, impressive as it may be, does not
straightforwardly tell us whether they have genuinely acquired
structured knowledge about syntax. We ought to consider the possibility
that they simply rely on recognizing shallow statistical patterns
observed in their enormous training data
\citep{millierePhilosophicalIntroductionLanguage2024}. These models are
undoubtedbly powerful statistical learners; but their capacity for
memorization and pattern matching could in principle explain their
ability to apply common grammatical constructions without assuming that
they acquire a deeper syntactic competence.

\begin{figure}

\centering{

\includegraphics{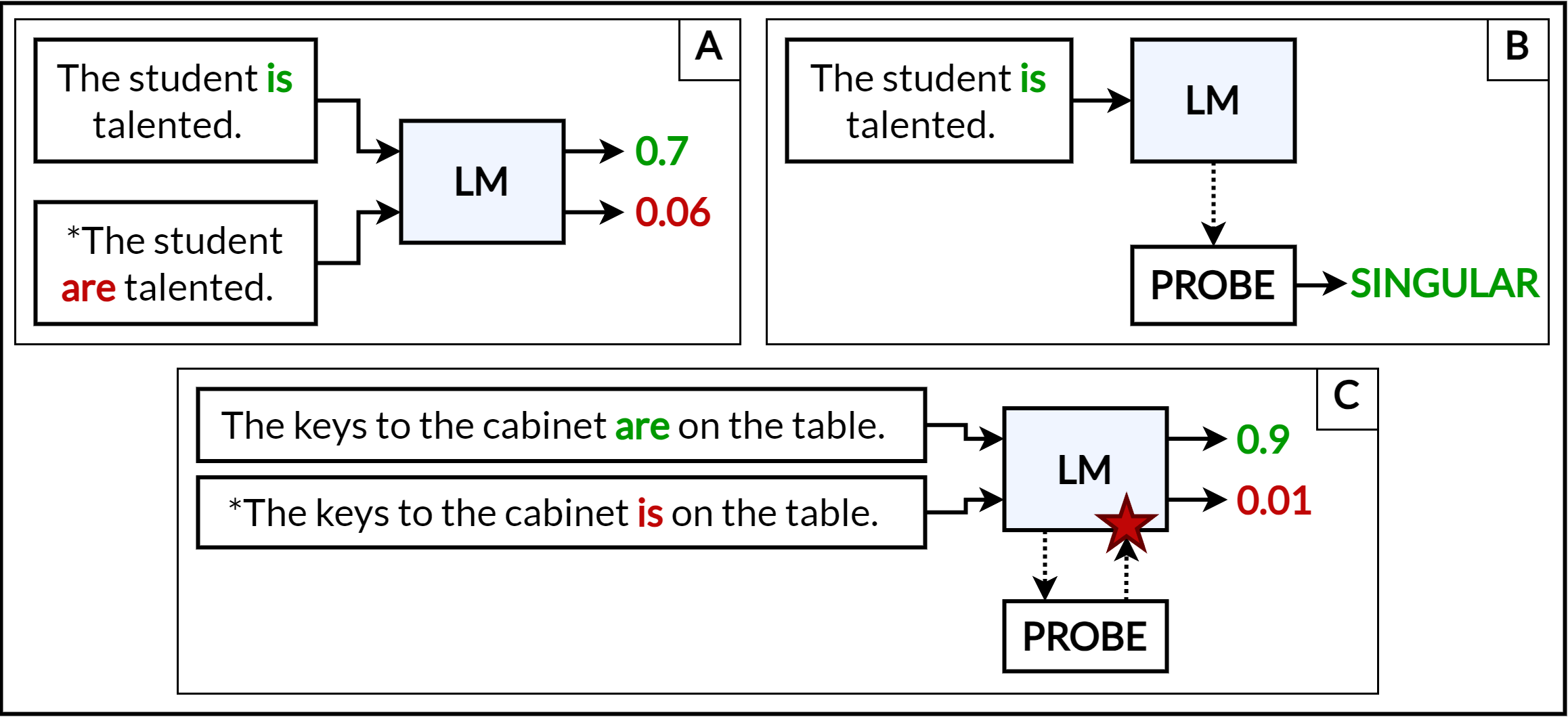}

}

\caption{\label{fig-methods}Three methodological approaches to assess
syntactic knowledge in language models}

\end{figure}

An increasingly large body of work in computational linguistics
investigates this question using complementary strategies. We can
distinguish three main methodological approaches: \emph{behavioural
studies} focus on models' responses to specific inputs
(Figure~\ref{fig-methods} A); \emph{probing studies} attempt to decode
information from models' internal activations (Figure~\ref{fig-methods}
B); and \emph{interventional studies} attempt to manipulate models'
internal states to determine how they causally influence behaviour
(Figure~\ref{fig-methods} C). These experimental strategies are largely
inspired from cognitive science -- particularly linguistics, psychology
and neuroscience -- but adapted to the meet the specific challenges and
opportunities of studying computational artifacts rather than human
subjects
\citep{frankBabyStepsEvaluating2023, millierePhilosophyCognitiveScience2024}.

\subsection{Behavioural studies}\label{behavioural-studies}

Behavioural studies focus on evaluating the linguistic abilities of
language models in controlled tasks targeting specific syntactic
phenomena. The goal of these studies is to assess which linguistic
features models are sensitive to, and whether their behaviour aligns
with human behaviour. This line of research takes a psycholinguistic
perspective, evaluating the implicit knowledge of neural networks
through experiments informed by human research -- what we might call
``linguistically oriented deep net analysis''
\citep{baroniProperRoleLinguistically2022}.

\subsubsection{Targeted syntactic tasks}\label{targeted-syntactic-tasks}

A common approach to behavioural experiments consists in selecting or
designing stimuli carefully chosen to exhibit a target linguistic
feature, and evaluating whether the target system is sensitive to that
feature. Linguists routinely use acceptability judgments to support
inferences about the grammaticality of particular constructions, under
the hypothesis that grammatical sentences tend to be judged as more
acceptable than ungrammatical ones
\citep{schutzeLinguisticEvidenceGrammatical2011, sprouseAcceptabilityJudgmentsGrammaticality2018}.
Accordingly, one may present sentences to a language model and assess
whether its behaviour aligns with human acceptability judgments. If the
model can reliably detect acceptable sentences in a given category, this
may be taken as evidence that it is sensitive to the corresponding
grammatical distinction.

Prompting language models to produce explicit grammaticality judgments
has yielded mixed results. \citet{dentellaSystematicTestingThree2023}
tested three variants of the language model GPT-3 on grammaticality
judgment tasks across eight linguistic phenomena, asking them directly
whether sentences were grammatically correct; they found above-chance
but low overall accuracy, greater accuracy for grammatical than
ungrammatical sentences, high response instability within items, and a
yes-response bias -- in contrast to human controls. On the other hand,
\citet{ambridgeLargeLanguageModels2024} prompted GPT-3 to rate the
grammatical acceptability of sentences on a 5-point scale, finding very
high correlation between the model's explicit acceptability ratings and
those of human adults for English causative sentences.

However, most behavioural studies do not prompt language models to
elicit explicit metalinguistic judgments, as this approach has been
shown to be unreliable and may lead to underestimating their actual
syntactic competence \citep{huPromptbasedMethodsMay2023}. For example,
answering a yes/no question about the grammaticality of a given sentence
requires not just grammatical competence, but also metalinguistic
knowledge of what grammaticality means, and the ability to verbalize
internal sensitivity to syntactic features. As such, this approach
imposes strong auxiliary task demands on language models that complicate
the interpretation of performance errors
\citep{huAuxiliaryTaskDemands2024}.

Instead, most behavioural experiments focus on minimal pairs of
sentences that only differ with respect to a specific syntactic
phenomenon -- such that one sentence is deemed grammatically acceptable
while the other is not -- and directly decode the probability assigned
by language models to a key word that differs across the pairs. Using
this methodology, \citet{huLanguageModelsAlign2024} show that
\citet{dentellaSystematicTestingThree2023}'s negative results should be
strongly qualified: language models evaluated on the same sentences in
minimal pairs achieve at- or near-ceiling performance on most linguistic
phenomena tested, except for centre embedding, where humans also perform
near chance; furthermore, minimal-pair surprisal differences strongly
predict human grammaticality judgments.\footnote{\citet{leivadaEvaluatingLanguageAbilities2024}
  and \citet{dentellaLanguageVivoVs2024} object to the use of direct
  probability measurements on the grounds that it does not allow for
  fair comparisons with human performance on acceptability judgment
  tasks. Indeed, we cannot obtain equivalent probability measurements
  from human subjects, and grammaticality for humans is -- on their view
  -- not a matter of relative comparison between sentences or of degree,
  but an absolute judgment about whether a sentence violates grammatical
  rules or not. However, comparing relative probabilities assigned to
  minimally different sentences does provide insight into models'
  sensitivity to specific syntactic features. While grammaticality may
  not be a matter of degree for individual sentences, the strength of
  preference between alternatives in a minimal pair can reveal graded
  aspects of linguistic knowledge that align with human judgments (as
  shown by \citet{huLanguageModelsAlign2024}).}

A good example of this strategy is the assessment of sensitivity to
subject-verb agreement, the phenomenon in which the form of a verb must
be congruent with the number and person of the subject in a sentence.
Subject-verb agreement in English arguably provides evidence for
hierarchical structure in syntactic processing, because the verb must
agree with the head of the subject phrase rather than the linearly
closest noun. Since language models process language sequentially
without built-in hierarchical representations of syntactic structure,
assessing their sensitivity to subject-verb agreement is an interesting
test of their general ability to learn syntactic rules. To rule out
alternative explanations of good model performance based on shallow
heuristics, such as mere sensitivity to linear order, it is common to
make minimal-pair tasks more challenging by including \emph{attractors}
in the stimuli. Attractors are chosen to have misleading features that
interfere with the surface properties of the sentence without actually
influencing its grammaticality (at least for idealized language users).

In a pioneering study, \citet{linzenAssessingAbilityLSTMs2016} tested
subject-verb agreement in an early language model based on a Long
short-term memory (LSTM) architecture rather than the Transformer. They
selected naturally occurring present-tense English sentences from
Wikipedia, including some sentences containing agreement attractors --
intervening nouns with a different number from the head subject noun.
For example, in the sentence ``The keys to the cabinet are on the
table,'' the plural noun phrase (``The keys'') governs the agreement
with the plural verb (``are''). However, the intervening singular noun
phrase (``the cabinet'') acts as an attractor; while syntactically
integrated into the subject within a prepositional phrase, it creates a
potential locality effect where the verb's proximity to a singular noun
might lead to confusion about subject-verb agreement. Linzen and
colleagues fed the selected sentences word-by-word into an LSTM language
model, comparing the probabilities assigned by the model to the two
forms of the focus verb (singular and plural). The verb form that the
model assigned the higher probability was selected as a proxy for
grammaticality judgments.

While supervised models trained with an explicit grammatical target
(e.g., number prediction or grammaticality judgments) achieved
near-perfect accuracy on simple cases with no attractors, the
unsupervised language models trained purely on next word prediction
faired worse (6.78\% error rate). The gap between supervised and
unsupervised models widened with the introduction of an increasing
number of attractors. Performance slowly degraded for supervised models,
only reaching an error rate of 17.6\% with four attractors; meanwhile,
the language model did much worse than chance in this most challenging
setup. Importantly, these initial results do not straightforwardly
translate to more modern architectures for language modelling. Indeed,
even small Transformer-based language models like BERT
\citep{devlinBERTPretrainingDeep2018} tested in another study performed
near-perfectly on the same task, with no noticeable performance
degradation on stimuli containing multiple attractors
\citep{goldbergAssessingBERTSyntactic2019}.

A follow-up study by \citet{gulordavaColorlessGreenRecurrent2018} set
out to control whether models might leverage semantic and
frequency-based cues rather than genuinely syntactic ones to achieve
good performance on tests of sensitivity to long-distance number
agreement. In addition to selecting a set of long-distance agreement
constructions from treebanks in several languages, they also created
``nonce'' versions of these test sentences by replacing all content
words with random words that have matching morphological features
(inspired by \citet{chomskySyntacticStructures1957}'s famous example
``Colorless green ideas sleep furiously''). This results in grammatical
but meaningless sentences that remove potentially helpful semantic and
frequency cues from the task. The results show that language models
based on a recurrent neural networks (RNN) architecture achieve high
accuracy on both original and nonce sentences, with only a small
reduction in accuracy for the latter. Their success on the nonce
sentences supports the conclusion that RNNs are acquiring useful
syntactic knowledge from language modelling, not just memorizing word
co-occurrence statistics. Additionally, the RNNs that perform best at
their language modelling objective (measured in terms of their
performance on next word prediction) also perform best on the agreement
task, providing further evidence of the relationship between language
modelling and syntactic knowledge. Once again, Transformer models like
BERT were found to perform better than RNNs on this task, including in
the nonce sentence condition \citep{goldbergAssessingBERTSyntactic2019}.

\citet{marvinTargetedSyntacticEvaluation2018} further tested LSTM
language models on a broader range of syntactic phenomena, including
subject-verb agreement, reflexive anaphora, and negative polarity items.
Crucially, they constructed minimal pairs of stimuli for each phenomenon
using templates instead of selecting naturally occurring sentences,
allowing them to achieve greater coverage and finer control of potential
confounds. They found that while LSTM models performed near-perfectly on
local subject-verb agreement dependencies, their performance degraded
substantially on non-local dependencies such as long VP coordination
(e.g., ``The manager writes in a journal every day and likes/*like to
watch television shows'') or agreement across a prepositional phrase
(e.g., ``The farmer near the parents smiles/*smile''). Likewise,
performance on reflexive anaphora (e.g.~``The manager that the
architects like doubted himself/*themselves'') and negative polarity
items (e.g.~``No/*Most students have ever lived here'') was mixed across
conditions, with models performing significantly worse than humans
overall. However, \citet{goldbergAssessingBERTSyntactic2019} also found
that Transformers performed much better, achieving near or above human
performance on these tasks.

\citet{wilcoxWhatRNNLanguage2018} evaluated models on filler-gap
dependencies. Filler-gap dependencies describe a syntactic construction
where a word or phrase (the filler), often a \emph{wh}-word like
``what'' or ``who,'' is moved to a different position in a sentence,
leaving behind an empty position (the gap), with both elements retaining
their semantic relationship within the sentence's structure. For
example, ``Who did you see \_\_ at the library?'' has a filler
(\emph{who}) moved to the front of the sentence, leaving a gap (marked
with underscores) after the verb. Without the filler, as in ``*You did
see \_\_ at the park'', the sentence becomes incomplete and incorrect.
Filler-gap dependencies are subject to complex island constraints:
specific syntactic environments where the usual relationship between the
filler and the gap is blocked, rendering certain configurations
ungrammatical \citep{rossConstraintsVariablesSyntax1967}. These
constraints delineate the boundaries within which the filler-gap
dependencies operate, such as prohibiting gaps within complex noun
phrases or in doubly nested clauses headed by \emph{wh}-words, thus
placing restrictions on where gaps can occur in a sentence. Wilcox and
colleagues found that LSTM language models are sensitive to filler--gap
dependencies and to some of the island constraints on them, in which
cases their expectation of a gap is attenuated. Whether these results
actually demonstrate model sensitivity to island constraints as opposed
to non-grammatical factors is debated
\citep{chowdhuryRNNSimulationsGrammaticality2018}, and additional
research suggests that RNNs are insensitive at least to some island
constraints \citep{chavesWhatDonRNN2020}.

Ruling out confounds from surface heuristics in targeted behavioural
studies is challenging, despite the use of experimental controls like
attractors. \citet{leeCanLanguageModels2022} investigated whether
language models could correctly predict agreement patterns between
reflexive pronouns (e.g.~``himself'') and their referent noun phrase in
English control constructions, which lack clear surface cues like
subject-verb agreement. They tested the Transformer-based language model
GPT-2 \citep{radfordLanguageModelsAre2019} on transitive control
constructions containing both a subject and object, with and without an
intervening noun phrase between the reflexive pronoun and its subject
controller. They found that GPT-2 performed at chance levels on subject
control constructions with an intervening noun, incorrectly relying on
agreeing with the closest noun phrase. However, the model performed at
ceiling on object control and on constructions without an intervening
noun between the reflexive and subject. Overall, the results suggests
that GPT-2's sensitivity to reflexive anaphor agreement patterns in
control constructions is limited, despite its strengths on other
syntactic tasks.

\citet{futrellNeuralLanguageModels2019} further asked whether the
behaviour of language models provides evidence that they can
incrementally represent syntactic state, in the way a symbolic
grammar-based model does using a stack-based parse. Through carefully
designed psycholinguistic experiments probing phenomena like garden path
effects and interpretation of subordinators, they showed that LSTMs can
implicitly capture aspects of hierarchical syntactic state from language
modelling objectives alone. However, they also suggest that fully
encoding the syntactic requirements of constructions may require
explicit syntactic supervision during training.

Beyond targeted behavioural studies, general benchmarks or ``challenge
sets'' covering a wide range of syntactic phenomena have been designed
to evaluate the performance of language models more holistically. One
such resource is BLiMP (Benchmark of Linguistic Minimal Pairs), a
large-scale benchmark testing 67 minimal pair types in English, each
comprising 1,000 pairs, organised into 12 broad categories spanning
morphological, syntactic, and semantic phenomena
\citep{warstadtBLiMPBenchmarkLinguistic2020}. GPT-2 was found to perform
best overall (81.5\% accuracy), although it still fell short of human
performance (88.6\% estimated individual human agreement based on
ratings on a forced-choice task). Another Transformer-based language
model, RoBERTa\textsubscript{BASE}
\citep{liuRoBERTaRobustlyOptimized2019b}, was found to achieve
near-human performance (within 2\% points of the human baseline or
better) on 6 out 12 BLiMP categories \citep{zhangWhenYouNeed2020}.
SyntaxGym is another holistic evaluation pipeline that streamlines the
evaluation of language models on standardized test suites targeting a
broad range of syntactic phenomena
\citep{gauthierSyntaxGymOnlinePlatform2020}. The larger version of GPT-2
(GPT-2-XL) achieved 89.97\% accuracy on test suites from SyntaxGym
\citep{huSystematicAssessmentSyntactic2020}.

Taken together, this body of evidence suggests that modern neural
networks trained on a language modelling objective, and especially those
based on the Transformer architecture, are sensitive to hierarchical
syntactic structure beyond surface heuristics. Indeed, their performance
on targeted behavioural tests appears to generalize fairly well to
previously unseen instances of many syntactic phenomena, including in
challenging cases involving attractors and long-range dependencies. In
many cases, model behaviour is in line with human performance on
grammaticality judgements. While the majority of the reviewed
experiments focus on the English language, this general trend appears to
hold across languages
\citep{ravfogelCanLSTMLearn2018, muellerCrossLinguisticSyntacticEvaluation2020, liAreTransformersModern2021, de-dios-floresComputationalPsycholinguisticEvaluation2022}.
The significance of these results should not be understated, as the
structure-dependent generalization exhibited by language models has
traditionally been assumed to require the kind of systematic
compositional rules found in symbolic parsers.

\subsubsection{Compositionality and
recursion}\label{compositionality-and-recursion}

Despite the apparent success of language models on a wide array of
behavioural experiments, there is an ongoing debate about whether their
performance is robust enough to warrant ascriptions of human-like
syntactic competence. In linguistics, the principle of compositionality
states that the meaning of a complex expression is determined by the
meanings of its constituent parts and the way in which they are
syntactically combined \citep{parteeMontagueGrammarMental1981}.
Compositionality is not just a property of linguistic expressions, but
also often considered an essential aspect of linguistic competence,
where it refers to the ability to systematically construct and
comprehend novel expressions by combining known meaningful elements
according to grammatical rules. This productive capacity for rule-based
combination is meant to explain how humans can generalize the production
and comprehension of an infinite number of sentences from a finite set
of words and rules, beyond memorized associations.

A longstanding critique of connectionist models is that they fail to
exhibit this ability, unlike their symbolic counterparts
\citep{fodorConnectionismCognitiveArchitecture1988, quilty-dunnBestGameTown2022}.
The proficiency of modern language models in processing and generating
seemingly novel sequences unseen in their training data has prompted an
effort to systematically evaluate their compositional aptitude
\citep[see][ for
reviews]{pavlickSemanticStructureDeep2022b, donatelliCompositionalityComputationalLinguistics2023}.
However, these models are typically trained on massive corpora
containing a huge variety of linguistic constructions. This makes it
difficult to discern whether they have truly learned the underlying
computational principles needed for systematic generalization, or
whether they are relying on having memorized a large inventory of
constructions during training
\citep{kimUncontrolledLexicalExposure2022}.

To better assess models' compositional abilities, research in
computational linguistics has turned to synthetic datasets specifically
designed to test compositional generalization in a controlled setting.
These datasets intentionally limit the constructions present during
training, and construct test sets requiring the composition of seen
components in new ways. For instance, the SCAN dataset contains a set of
natural language commands (e.g., ``jump twice'') mapped to sequences of
actions (e.g., JUMP JUMP), with the test set containing longer commands
requiring systematic composition
\citep{lakeGeneralizationSystematicityCompositional2018}. Other
prominent examples are the CFQ dataset, which maps natural language
questions to logical forms
\citep{keysersMeasuringCompositionalGeneralization2019}, and the COGS
dataset, which tests generalization to unseen syntactic structures
\citep{kimCOGSCompositionalGeneralization2020}. By training models on
synthetic data, the aim is to evaluate whether they can productively
combine known units based on a representation of their underlying
structure, rather than relying solely on memorized patterns. This method
is reminiscent of ``control rearing'' studies in animal cognition
research, which also involve manipulating the learning environment of a
subject to evaluate its influence of a target behaviour
\citep{frankBabyStepsEvaluating2023}.

Initial results of testing deep neural networks on synthetic datasets
for compositional generalization generally showed a performance gap
between the training and test sets. This was suggestive of a limited
ability to properly generalize across challenging distribution shifts
that require productive combination of known elements in novel ways.
However, since then, many Transformer-based models have achieved strong
accuracy on compositional generalization datasets. This progress has
been enabled by various strategies, including modifications to the
standard Transformer architecture to provide more effective inductive
biases for compositionality
\citep{csordasDevilDetailSimple2021, ontanonMakingTransformersSolve2022},
and data augmentation techniques to expose models to a greater diversity
of training examples
\citep{andreasGoodEnoughCompositionalData2020, akyurekLearningRecombineResample2020, akyurekLexSymCompositionalityLexical2023, qiuImprovingCompositionalGeneralization2022a}.

Another promising strategy that has shown excellent results without
requiring architectural changes is \emph{meta-learning}, or learning to
learn better by generalizing from exposure to many related learning
tasks
\citep{lakeCompositionalGeneralizationMeta2019, conklinMetaLearningCompositionallyGeneralize2021, lakeHumanlikeSystematicGeneralization2023}.
Standard supervised learning relies on the assumption that training and
test data come from the same distribution, which can lead models to
overfit on the peculiarities of the training set. Meta-learning exposes
models to a distribution of related tasks, rather than a single task, to
promote learning of generalizable knowledge that transfers better. This
makes models less prone to memorizing training data, and better able to
productively combine known elements in new ways when faced with novel
combinations unseen during training. There is also evidence that
generalization accuracy on syntactically novel items from the
out-distribution test sets improves long after in-domain validation
accuracy on the training distribution plateaus; this suggests that
halting training too early based on in-domain validation accuracy leads
to greatly underestimating the ability of Transformer models to
generalize \citep{murtyGrokkingHierarchicalStructure2023}.

Another core tenet of theoretical linguistics holds that human
linguistic competence is linked to the ability for recursive processing.
Humans can construct and compute over hierarchically nested syntactic
representations by recursively applying functions to their own outputs,
with clauses embedded within other clauses in complex tree structures.
The processing of such recursive embeddings is taken to be a hallmark of
the human language faculty in the Chomskyan tradition, allowing for the
generation of syntactic structures with potentially unlimited complexity
from finite means
\citep{chomskySyntacticStructures1957, hauserFacultyLanguageWhat2002}.
Accordingly, assessing whether language models' ability to handle
recursion is deeply relevant to the discussion of their putative
syntactic competence beyond shallow pattern recognition.\footnote{It is
  worth noting that the assumption that recursion is a necessary
  property of all human languages has been challenged. For examples,
  languages like Riau Indonesian provide some evidence that linear
  grammars without recursion may be possible
  \citep{gilRiauIndonesianPivotless1999, nefdtPhilosophyTheoreticalLinguistics2024}.}

RNNs -- and their LSTM variants -- process data recurrently by applying
the same weights to sequential elements, maintaining a hidden state that
carries information across steps. In a classic study,
\citet{elmanDistributedRepresentationsSimple1991} showed that simple
RNNs trained on sentences containing multiply-embedded relative clauses
could encode information about their recursive structure, inspiring
research on connectionist models of recursive processing in humans
\citep{christiansenConnectionistModelRecursion1999}. By contrast with
RNNs, modern Transformers do not have built-in recurrent processing;
their self-attention mechanism endows them with distinct inductive
biases that lead them to process recursive constructions differently,
but may in fact give them an advantage in handling hierarchical
structure, as evidenced by their superior performance across a broad
range of complex syntactic tasks.

Recent research sheds light on the strengths and weaknesses of different
language models architectures when it comes to processing recursion.
\citet{lakretzMechanismsHandlingNested2021} found that while LSTM-based
language models can track information about local and long-distance
number agreement, they have a limited capacity to handle nested
recursive structures, seen in their failure to track agreement in some
long-range embedded dependencies. In a follow-up study,
\citet{lakretzCanTransformersProcess2022} investigated whether the newer
Transformer architecture shows improvements in processing dependencies
in nested constructions and can approximate human recursive competence.
They found that Transformer models like GPT-2-XL could process
short-range recursion in nested object-relative clauses nearly
perfectly, vastly exceeding LSTMs. However, their performance sharply
dropped below chance after adding a three-word prepositional phrase to
make the embedded dependency longer (e.g., ``The keys that the man
\textbf{near the cabinet} holds are\ldots{}'').\footnote{The much larger
  model GPT-3 \citep{brownLanguageModelsAre2020} also failed to perform
  above chance in the same condition, tested in the
  \texttt{subject\_verb\_agreement} subtask of the BIG-Bench benchmark
  \citep{srivastavaImitationGameQuantifying2023}.} They conclude that
Transformer-based models are fundamentally limited in their capacity to
handle long-range recursive nesting, and thus fail to model a core
aspect of human linguistic competence.

However, a closer look at the methodology of these studies suggests that
initial results should be interpreted with caution. Indeed, human
subjects tested by \citet{lakretzMechanismsHandlingNested2021} on
subject-verb agreement in nested sentences with centre embedding
received substantial training with examples, instructions, and feedback.
By contrast, neural networks were tested in a zero-shot setting, without
examples or task context. To assess the influence of this discrepancy,
\citet{lampinenCanLanguageModels2023} tested the Transformer-based
language model Chinchilla
\citep{hoffmannTrainingComputeOptimalLarge2022} on the same task,
providing it with context analogous to human training. When prompted
with several example sentences before each test case, Chinchilla
performed \emph{better} than humans even on the most challenging
conditions. Furthermore, upon reanalysing the human results of
\citet{lakretzMechanismsHandlingNested2021} on the task, Lampinen found
that human subjects, even after training, seem to perform near chance
the first few times they encounter difficult syntactic structures. These
results suggest that Transformer-based language models can in fact
handle complex nested syntactic dependencies as well as humans, given
just a few prompting examples, and that humans may also need some
experience on the task before performing well on the most complex cases.

More generally, this work highlights the difficulty of establishing fair
and meaningful behavioural comparisons between the behaviour of humans
and language models in an experimental context. Differences in task
framing can obscure real similarities or differences in the capabilities
being studied. When neural networks appear to exhibit performance
failures compared to humans, care should be taken to ensure that
experimental conditions are well-matched and take into account
contingent constraints on performance for all tested subjects or systems
\citep{firestonePerformanceVsCompetence2020}. This methodological
concern is familiar from comparative and developmental psychology, where
infants and non-human animals are known to exhibit specific performance
constraints like limited memory or motor control that can prevent them
from demonstrating full competence through behaviour on a task
\citep{frankBabyStepsEvaluating2023}. The discrepancy between language
models' performance on explicit grammaticality judgment tasks and direct
probability assessments is indicative of an analogous effect of
auxiliary factors related to the strength of task demands
\citep{huAuxiliaryTaskDemands2024, milliereAnthropocentricBiasPossibility2024}.
This point cuts both ways, however; the mere fact that language models
achieve high accuracy on a syntactic task or benchmark does not
straightforwardly entail that they possess the corresponding competence,
if potential confounds such as shallow heuristics are not adequately
controlled.

\subsection{Probing studies}\label{probing-studies}

While behavioural experiments can provide evidence regarding the
sensitivity of model predictions to syntactic phenomena in carefully
controlled conditions, they generally do not warrant stronger inferences
about how models represent this information internally. Probing studies
aim to go beyond mere behavioural data to determine what kind of
linguistic information can be extracted from the internal
representations of language models tested on specific tasks
\citep{alainUnderstandingIntermediateLayers2018, adiFinegrainedAnalysisSentence2016, shiDoesStringBasedNeural2016, hupkesVisualisationDiagnosticClassifiers2018}.

\subsubsection{Diagnostic probing}\label{diagnostic-probing}

The typical methodology of probing studies involves training a separate
supervised classifier, also called a \emph{diagnostic probe}, to predict
linguistic properties like part-of-speech tags or dependency relations
directly from the model's internal activations.\footnote{While
  behavioural studies are inspired by psycholinguistics, probing studies
  are inspired by decoding methods in neuroscience
  \citep{ivanovaProbingArtificialNeural2021}. By training classifiers on
  known patterns of brain activity associated with specific stimuli or
  tasks, neuroscientists can then use these classifiers to predict or
  decode the stimuli or task from new patterns of brain activity.
  Probing an artificial neural network offers much more granular and
  direct access to internal states than available with current
  neuroimaging techniques.} This generally involves collecting a set of
samples labelled with the target linguistic property, feeding these
samples as input to the model, and capturing the model's activations in
a given layer in response to each input. The resulting dataset of
activations-label pairs can be used to train the probing classifier to
predict each sample's label from the corresponding model activations. If
the probe can predict the target linguistic properties with high
accuracy on held-out examples, it suggests that information about those
properties is encoded in the model's learned representations. For
example, given a corpus of sentences labelled with part-of-speech tags
for each word, a linear classifier could be trained to map a model's
word embeddings to the correct part-of-speech tag. After training, the
probe's performance at assigning the correct part-of-speech tag to new
unseen words would be evaluated. High accuracy on the test set suggests
the model's word embeddings encode information relevant for
part-of-speech disambiguation.

Probing classifiers have been widely applied to study many model
architectures and linguistic phenomena \citep[see][ for a
review]{belinkovProbingClassifiersPromises2022}. Early examples include
probing a neural machine translation model to predict morphological
properties \citep{shiDoesStringBasedNeural2016}, probing an LSTM tested
on a subject-verb agreement task to decode information about the
subject's number \citep{giulianelliHoodUsingDiagnostic2018}, and probing
sentence embeddings to examine which syntactic properties such as parse
tree depth are encoded \citep{conneauWhatYouCan2018}. More recent work
has scaled up probing to analyse larger pre-trained Transformer language
models, with a particular focus on BERT due to its wide availability.
This research program, informally known as ``BERTology''
\citep{rogersPrimerBERTologyWhat2020}, yielded converging evidence that
a broad range of syntactic information is decodable from the internal
activations of language models. This includes evidence that BERT encodes
hierarchical rather than merely linear structure
\citep{linOpenSesameGetting2019, warstadtCanNeuralNetworks2020};
evidence that BERT's contextual word embeddings encode information about
part-of-speech tags, syntactic chunks, dependencies, semantic roles and
coreferents hierarchically organized across layers
\citetext{\citealp{tenneyBERTRediscoversClassical2019}; \citealp{liuLinguisticKnowledgeTransferability2019}; \citealp{jawaharWhatDoesBERT2019}; \citealp[although
see][ for a critical discussion of the relationship between layer depth
and decodable information]{niuDoesBERTRediscover2022}}; and even
evidence that syntactic parse trees can be recovered from BERT and its
variants
\citep{vilaresParsingPretraining2020, kimArePretrainedLanguage2019, rosaInducingSyntacticTrees2019, arpsProbingConstituencyStructure2022}.

\subsubsection{Methodological
challenges}\label{methodological-challenges}

Standard probing studies using diagnostic classifiers have raised
methodological concerns \citep{belinkovProbingClassifiersPromises2022}.
The main concern is that probing is fundamentally correlational -- the
presence of decodable information does not conclusively demonstrate it
plays a causal role in the model's outputs. When a probe achieves high
accuracy on the prediction of a linguistic feature from a model's
activations, this may not straightforwardly entail that the model
actually represent the relevant feature without adequate control. Two
primary alternative explanations of probe accuracy must be ruled out to
support claims about linguistic competence.

The first possibility is that, while the probed linguistic feature is
decodable from the model's activations, the model does not in fact
utilize this information when making predictions. In other words, the
classifier's high accuracy in predicting a feature from the model's
activations may indicate that while the latter genuinely encodes the
relevant information, it is not causally efficacious in model behaviour
on the task. There is empirical evidence that this concern is not always
unfounded. For example, \citet{elazarAmnesicProbingbehavioral2021} used
a technique called adversarial probing to explicitly remove information
about specific linguistic features (e.g., part-of-speech information)
from a model's activations. By measuring the subsequent impact on the
model's core task performance, they demonstrated that high decoding
accuracy does not always demonstrate a major causal effect of decoded
information.

The second possibility is that the relevant linguistic knowledge is not
even encoded in the model's learned representations; rather, the probing
classifier might be powerful enough to recover the property from surface
patterns, memorization, or other cues
\citep{hewittDesigningInterpretingProbes2019}. This is because probes
are trained in a supervised manner on task data labelled for the
linguistic property of interest. As a result, the probe has access to
explicit supervision teaching it to recognize patterns related to the
property in model activations, and may learn from irrelevant
correlations. For instance, nonlinear classifiers may be able to predict
syntactic properties by memorizing the training set instead of
extracting syntactic features from the representation. This risks
wrongly ascribing complex linguistic capabilities to the model when
probing does not convincingly demonstrate such knowledge was already
present before introducing the explicit probing dataset.

Control tasks and metrics such as selectivity can help distinguish
between genuine extraction of linguistic structure and reconstruction
from spurious cues. For example,
\citet{hewittDesigningInterpretingProbes2019} designed control tasks
where the labels were randomly shuffled, so the probe could only succeed
by memorizing spurious cues rather than extracting linguistic knowledge.
They defined selectivity as the gap between probing accuracy on the real
task versus the control task. High selectivity suggests that the probe
is genuinely extracting linguistic properties from the model's
activations, while low selectivity suggests it may be latching onto
spuriously predictive cues.

There is an ongoing debate about whether simple linear probes should be
used to minimize the risk of learning to reconstruct linguistic features
from spurious cues. Linear probes have been advocated under the
assumption they have lower expressive power compared to nonlinear
probes. As such, they are more likely to rely solely on features
explicitly encoded in the model's activations, while more complex probes
are more likely to capture additional signals that are not actually used
by the model itself. However, the potential trade-off between probe
complexity and accuracy may warrant the use of complex probes if proper
controls are in place. In particular, more complex probes may provide a
less constrained estimate of the total discriminative information about
a property encoded in the model's activations
\citep{pimentelParetoProbingTrading2020}.

\subsubsection{Parameter-free probing}\label{parameter-free-probing}

Parameter-free approaches constitute an interesting alternative to using
supervised classifiers. The goal of parameter-free probing is to extract
information about a model's encoding of a linguistic feature by directly
analysing its representations without introducing additional learned
parameters that may confound the results. One such strategy consists in
analysing the self-attention weights of Transformer-based language
models to recover syntactic information. Attention heads were found to
track dependencies (e.g.~objects of verbs, determiners of nouns,
prepositional objects, and coreference) with high accuracy
\citep{raganatoAnalysisEncoderRepresentations2018, clarkWhatDoesBERT2019, marecekBalustradesPierreVinken2019},
with some attention heads specializing in tracking individual dependency
types \citep{htutAttentionHeadsBERT2019}.
\citet{cherniavskiiAcceptabilityJudgementsExamining2022} used
topological data analysis to extract graph-based features from
Transformer attention maps. They showed topological properties of the
attention graph improve acceptability classification and minimal pair
detection without additional parameters, revealing interpretable
correlations between attention patterns and specific grammatical
phenomena.

Representational similarity analysis (RSA) is an alternative
parameter-free method from computational neuroscience
\citep{kriegeskorteRepresentationalSimilarityAnalysis2008} that involves
measuring the similarity of model representations to prototypical
representations constructed to instantiate specific linguistic
hypotheses. Using RSA, \citet{leporiPickingBERTBrain2020} found that the
representational geometry of BERT's contextual word embeddings reflects
specific syntactic dependencies (e.g.~pronoun coreference and verb
subject-sensitivity) better than random controls. Similarly,
\citet{chrupalaCorrelatingNeuralSymbolic2019} used RSA to find a
significant correspondence between the representational geometry of
various language models and a reference model based on gold syntax
trees.

\citet{wuPerturbedMaskingParameterfree2020} proposed another
parameter-free technique called perturbed masking, which masks different
words and analyses impact on model predictions. They derived a word
impact matrix from which they extracted unlabelled dependency trees with
high accuracy; furthermore, the induced dependency trees improved model
performance on downstream tasks (e.g., sentiment analysis) despite
differences from human-designed parsers.

Finally, \citet{murtyCharacterizingIntrinsicCompositionality2022}
recently introduced an interesting parameter-free method called
\emph{tree projection} to probe the intrinsic compositionality of
Transformer models. Tree projection measures the ``tree-structuredness''
of the model's internal computations on an input, by scoring how well
they can be approximated by explicitly tree-structured models. The
authors found that Transformers trained on compositional generalization
datasets become increasingly tree-like over the course of training, with
tree projections progressively matching ground truth syntax.
Tree-structuredness also positively correlates with compositional
generalization. The emergence of tree-like computation and alignment
with syntactic formalisms provides evidence that Transformers can learn
to implicitly encode hierarchical syntactic knowledge, despite lacking
explicit architectural constraints for tree-like structures.

\subsection{Interventional studies}\label{sec-interventional-studies}

While parameter-free probing avoids potential confounds introduced by
training an additional supervised classifier on top of the model's
representations, it remains, like diagnostic probing, fundamentally
correlational. Demonstrating that a linguistic feature can be decoded
through attention weights, similarities to prototypes, or tree
projections does not guarantee that information plays a causal role in
model predictions. As such, mere probing provides an upper bound on
relevant information that the model could use. By contrast,
interventional studies aim to demonstrate causal efficacy by actively
interfering with the model's internal states to assess their impact on
behaviour.

\subsubsection{Counterfactual
interventions}\label{counterfactual-interventions}

\citet{giulianelliHoodUsingDiagnostic2018} offer a classic example of
this interventional approach. After training a diagnostic probe to
predict number agreement from the activations of an LSTM language model,
the authors actively intervened on the activation pattern identified by
the probe to influence the behaviour of the model. Specifically, in
cases where the model failed on
\citet{gulordavaColorlessGreenRecurrent2018}'s subject-verb agreement
prediction task, they modified the relevant activation pattern in the
model such that the diagnostic classifier's agreement prediction would
move slightly closer to the ground truth. This causal intervention
successfully improved the accuracy of the model on the task, providing
evidence that the model's encoding of subject-verb number agreement
information is causally efficacious.

\citet{ravfogelCounterfactualInterventionsReveal2021} introduced a novel
causal probing method called AlterRep that involves generating
counterfactual representations by manipulating the model's encoding of
specific linguistic features. They applied it to assess whether BERT
leverages relative clause boundary information encoded in its
activations correctly when predicting subject-verb number agreement in
English. By manipulating the model's encoding of whether words are
inside or outside relative clauses, they found that BERT's agreement
predictions changed systematically in alignment with proper relative
clause usage. This suggests that BERT does use relative clause boundary
information in a causal, generalizable way for grammatically correct
number agreement, consistently with the rules of English grammar.

\citet{lasriProbingUsageGrammatical2022} took a usage-based approach to
probing how linguistic properties are functionally encoded in language
models. As a case study, they focused on grammatical number and its role
in subject-verb number agreement. After confirming that number is
encoded in BERT's embeddings, they performed causal interventions to
erase number information at different layers and analyse impacts on
agreement accuracy. This approach builds on amnesic probing
\citep{elazarAmnesicProbingbehavioral2021}, but is tailored to
linguistic behavioural tasks requiring the erased information
\citep{linzenAssessingAbilityLSTMs2016}. The precise alignment between
lost number information and degraded agreement performance provides
strong evidence that BERT relies on the erased encodings to perform
number agreement. Further experiments revealed that BERT employs
distinct subspaces for encoding number in nouns versus verbs, with
information transferred indirectly across intermediate layers. Using
counterfactual interventions, \citet{haoVerbConjugationTransformers2023}
also show that BERT's ability to conjugate verbs is determined by linear
encodings of subject number that are distributed across token positions
in middle layers and concentrated in the subject position in early
layers and verb position in later layers.

These causal interventions go a long way towards establishing the claim
that language models do represent syntactic features. In accordance with
prominent philosophical theories of representation, for a pattern of
activation \(A\) in a model to represent a given feature \(F\) in the
context of a given task, it must not only be the case that \(A\) bears
correlational information about \(F\), but also that the system uses the
relevant information to succeed at the task, \emph{and} that it can
misrepresent \(F\) with some inputs
\citep{sheaRepresentationCognitiveScience2018}. The existence of a probe
that successfully predicts \(F\) from \(A\) provides evidence for the
first condition (correlational information); task performance
degradation following a causal intervention to reduce information about
\(F\) in the model provides evidence for the second condition (usage of
information); finally, task performance improvement following a causal
interventions to shift probe prediction closer to the true labels
provides evidence for the third condition (misrepresentation)
\citep[see][ for a detailed
discussion]{hardingOperationalisingRepresentationNatural2023}.
Accordingly, causal probing studies do provide preliminary evidence for
representational claims about syntactic features in language models.

\subsubsection{Mechanistic
interpretability}\label{mechanistic-interpretability}

While causal probing provides targeted evidence about the encoding of
linguistic features, the nascent field of mechanistic interpretability
takes a more comprehensive approach to reverse engineering models'
internal computations
\citep[@millierePhilosophicalIntroductionLanguage2024a]{elhage2021mathematical}.
This research programme builds on a loose analogy between neural
networks and traditional computer programs, proposing that we might
rigorously ``reverse engineer'' neural networks to recover
human-interpretable descriptions of how they process information, akin
to decompiling software. At the core of this paradigm is the notion that
neural networks represent information in terms of interpretable features
connected through learned ``circuits'' or subnetworks implementing
meaningful computations \citep{olahZoomIntroductionCircuits2020}. For
example, a circuit might route semantic information between embeddings
across different layers based on syntactic relationships. By formally
characterizing such circuits in terms of weights and activations,
mechanistic interpretability aims to provide complete functional
explanations of model behaviour.

To make progress towards this reverse-engineering goal, mechanistic
interpretability researchers use interventions analogous to causal
probing techniques, which involve deleting or replacing pieces of a
model to identify components critical for certain computations. For
example, \citet{wangInterpretabilityWildCircuit2022} discovered a
circuit for indirect object identification in GPT-2-small using a
combination of interpretability approaches relying on causal
interventions. More sophisticated approaches can also be used to
investigate the representation of syntactic knowledge at a finer level.
A promising recent development involves training sparse autoencoders
(SAEs) to identify interpretable features within the model's hidden
states in an unsupervised manner without relying on ad hoc probes. For
example, \citet{marksSparseFeatureCircuits2024} trained SAEs on the
activations of a small language model (Pythia-70M) and used integrated
gradients to compute approximate indirect effects of SAE features on the
model's output for contrastive pairs of sentences. This process allows
for the discovery of ``sparse feature circuits'' that reveal the model's
internal mechanisms for performing linguistic tasks. They found that
small feature circuits of fewer than 100 nodes can explain a large
proportion of the model's behaviour in subject-verb agreement tasks.
Specifically, they identified an interpretable algorithm that detects
the main subject's grammatical number in early layers, identifies
distractors like the start of a relative clause or prepositional phrase,
and moves the subject number information to the end of the distractor
clause, such that it can be used in the model's final layers to promote
matching verb forms. Notably, the circuits for handling simple agreement
and different intervening clauses showed substantial overlap, suggesting
that this small model had developed a relatively abstract and general
mechanism for subject-verb agreement.

Going beyond the investigation of grammatical rules,
\citet{yamakoshiCausalInterventionsExpose2023} used causal interventions
to analyse how language models process Winograd Schema Challenge
sentences, which require commonsense reasoning to resolve ambiguous
pronouns. Their found distinct circuits within the model responsible for
integrating contextual information, suggesting that language models may
construct implicit ``situation models'' to resolve ambiguities.

These techniques can also be used to study the learning dynamics of
neural networks. Studying a small Transformer model trained on a modular
addition task through the lens of mechanistic interpretability,
\citet{nandaProgressMeasuresGrokking2022} found evidence of three
distinct learning phases: an initial phase in which the model relies on
brute memorization, an intermediate phase in which it forms a dedicated
circuit implementing a general algorithm for modular addition, and a
cleanup phase in which the memorization components are removed. This
raises the intriguing possibility that a similar learning process could
occur for the acquisition of syntactic rules when a model is trained on
natural language data: over the course of training, language models
might be forced to learn syntax to improve their performance on
next-word prediction, after an initially relying on memorizing
constructions \citep{murtyGrokkingHierarchicalStructure2023}. In fact,
\citet{chenSuddenDropsLoss2023} identified a syntax acquisition phase in
the training of masked language models, characterized by sudden drops in
loss and rapid improvements in syntactic capabilities. They observed two
distinct phase transitions: a ``structure onset'' marked by a spike in
unlabelled dependency parsing accuracy, followed by a ``capabilities
onset'' where the model shows an abrupt increase in performance on the
BLiMP grammatical acceptability benchmark.

While there isn't much overlap yet between research on causal probing in
computational linguistics and research on mechanistic interpretability,
the latter could help answer questions about whether and how language
models implement syntactic rules at a much finer level of granularity.
Having reviewed empirical evidence for syntactic knowledge in language
models, I will now consider the implications of these findings for
theoretical linguistics and ongoing debates about language acquisition.

\section{Language models and theoretical
linguistics}\label{sec-language-models-linguistics}

Experimental research on language models is almost completely ignored in
theoretical linguistics \citep[see][ for a quantitative analysis of the
literature]{baroniProperRoleLinguistically2022}. The reverse is not
true, as many of these experiments in computational linguistics are
explicitly informed by linguistic theory. This asymmetry calls for an
explanation. It could be that most theoretical linguists are not
well-acquainted with the experimental literature on language models; or,
on a more charitable view, they might think this literature is
irrelevant to their own theoretical projects. The latter explanation is
certainly true of some vocal critics of language models: Noam Chomsky,
for example, has prominently argued that there is nothing
\emph{whatsoever} they could contribute to linguistics even in principle
\citetext{\citealp{chomskyNoamChomskyFalse2023}; \citealp[see
also][]{norvigChomskyTwoCultures2017a}}.\footnote{``In principle, they [language models] can tell us nothing about language, language acquisition, human cognition, anything" (Chomsky, personal communication).}
If this were true, then the whole body of research discussed in the
previous section would have nothing to tell us about human language use
and acquisition, and could be seen as a mere exercise in studying
engineering artifacts.

If we take language models seriously as \emph{models}, rather than mere
engineering artifacts, what are they models \emph{of}? The answer to
this question informs the relevance of language models to linguistics.
In what follows, I will discuss three modelling targets for language
models: \emph{linguistic performance}, \emph{linguistic competence}, and
\emph{language acquisition}. These possibilities are not mutually
exclusive. Language models may differ in various ways, including
architecture, parameter size, and training data; these differences are
relevant to what a given model can reasonably be taken to be modelling.
Furthermore, it is conceivable that the same language model might be
used to investigate aspects of linguistic performance, competence, and
acquisition, depending on the pragmatic goals of researchers in
particular experimental contexts. This still leaves open the question
whether language models could, \emph{in principle}, be treated as models
of performance, competence, or acquisition. If the answer is negative,
particularly when it comes to modelling linguistic competence and
acquisition, then the relevance of language models to theoretical
linguistics ought to remain very limited. If the answer is positive,
however, then these models might be very useful tools, in the right
experimental context, to test linguistic hypotheses and constrain
linguistic theorizing.

\subsection{Performance and
competence}\label{performance-and-competence}

In mainstream generative linguistic theory, a key distinction is made
between a speaker's linguistic \emph{competence} -- their idealized
knowledge of a language's grammar -- and their \emph{performance} -- the
constrained manifestation of this competence in actual language use
\citep{chomskyAspectsTheorySyntax1965}. On this view, publicly
observable utterances result from unobservable internal structures, but
also involve many additional psychological processes beyond core
competence. Linguistic performance can be affected by external factors
like memory limitations, distractions, slips of the tongue, etc. that
may obscure the full extent of the underlying competence.

What would it mean for a language model to be a model of human
linguistic performance, as opposed to competence? An obvious proposal is
that this is merely a matter of behaviour: if the language model
\emph{behaves} similarly enough to humans in a broad range of linguistic
scenarios -- for example by matching human response patterns on
sophisticated benchmarks probing various aspects of linguistic
performance --, then it is \emph{ipso facto} a predictive model of human
linguistic performance.\footnote{In Section~\ref{sec-scientific-models},
  I will come back to the notion of scientific model in philosophy of
  science, and the extent to which it applies to language models
  depending on what they are intended to be modeling.} Given that
language models are trained on human-generated text corpora with a
next-word prediction objective, all sufficiently trained language models
can be treated as models of linguistic performance in this sense. In
fact, language models trained on a large amount of data do excel at
mimicking the (written) outputs of human language users, which is why it
has become particularly challenging or even impossible to automatically
detect machine-generated text \citep{sadasivanCanAIGeneratedText2023}.
This does not entail, however, that all language models are equally
\emph{good} models of performance. Larger language models like GPT-4
perform better than smaller models at next-word prediction (measured by
\emph{perplexity}, a metric that quantifies how well the probability
distribution predicted by the model aligns with the actual distribution
of the words in the text). But they also hardly ever make grammatical
mistakes -- unlike humans. In that respect, it is unclear that they
should be treated as the \emph{best} models of human performance,
compared to less grammatically proficient models.

Furthermore, linguistic performance does not merely encompass language
production, but also language comprehension. Expectation-based theories
of sentence processing posit that processing difficulty is driven by the
predictability or surprisal of upcoming words based on the context
\citep{levyExpectationbasedSyntacticComprehension2008}. In
psycholinguistics, reading times are routinely used as a proxy measure
for the predictability of upcoming words, as reflected in the subjective
probabilistic expectations humans form during language processing.
Interestingly, language models' performance on next-word prediction
(perplexity) appears to be correlated to their ability to predict human
reading times up to a certain point
\citep{wilcoxPredictivePowerNeural2020}. However,
\citet{shainLargeScaleEvidenceLogarithmic2022} found that GPT-2-small
significantly outperformed larger models, like GPT-3, in predicting
human reading times across a number of different datasets -- despite the
fact that larger models are better at next-word prediction \citep[see
also][]{ohTransformerBasedLMSurprisal2023}. This suggest that there is a
threshold beyond which next-word prediction performance no longer
reflects human subjective word probabilities. This is noteworthy given
that smaller models like GPT-2-small are trained on a more realistic
quantity of linguistic data, compared to the linguistic input of humans
(see Section~\ref{sec-model-learners} below).

The question whether (some) language models might be treated as models
of linguistic \emph{competence} is more difficult, and crucially hinges
on controversial theoretical assumptions. If performance closely
reflects competence, then modelling performance with a language model
trained on linguistic utterances could in principle provide insight into
human competence. This assumes that the effects of performance are
merely due to noise in the externalization of competence through
behaviour, and do not systematically prevent inference of the full
underlying competence structure. However, generative linguists deny this
assumption, pointing to para-linguistic performance effects like
interjections and hypothetical complex transformations undergone by
linguistic structures in the process of externalization. To get at these
hypotheses about competence, theoretical linguistics draws on various
sources of evidence, including evolutionary theory and developmental
psychology, rather than merely modelling the surface forms of language.

\citet{dupreWhatCanDeep2021} argues that if there is indeed a
substantial gap between human linguistic competence and performance,
then training a language model to mimic linguistic performance through
next-word prediction may tell us little about the competence that
theoretical linguists aim to describe. This would entail that
computational linguistics research on language models is mostly
irrelevant to theoretical linguistics, particularly to discriminate
between competing theories of linguistic competence. Even if a model
achieves human-like performance on a syntactic task, and even if we
manage to infer the computations underlying such behaviour through
causal interventions, we cannot directly infer that human performance on
same task is underlain by the same computations.

Note that this argument does not entail that language models have
nothing to tell us about language acquisition. As we will see in what
follows, experiments with language models may be very relevant not only
to theoretical learnability claims, but also to weaker developmental
claims. In addition, Dupre doesn't deny that neural network models may
constrain linguistic theories at least indirectly, by offering some
evidence about the developmental or neurobiological plausibility of some
proposed competences compared to others
\citep[619-20]{dupreWhatCanDeep2021}. Finally, he concedes that language
models \emph{could}, in principle, acquire human-like competence rules
merely from being trained on performance data; however, he suggests that
this would be surprising given the systematic gaps between competence
and performance, and the diverging goals of theoretical linguistics and
language modelling (explanation vs.~prediction).\footnote{An alternative
  suggestion is that language models should be viewed as
  implementation-level models of linguistic competence, representing a
  mechanistic abstraction of neural processes that implement linguistic
  computations \citep{blankWhatAreLarge2023}. This interpretation is
  supported by studies mapping the activations of language models to
  brain signals and by findings that language models can predict neural
  responses to language across various neuroimaging modalities
  \citep[see][ for a review]{tuckuteLanguageBrainsMinds2024}. I will
  leave that suggestion aside here to focus on the direct relevance of
  language models to linguistic theory.}

Dupre's argument is ultimately conditional: to the extent that
mainstream generative linguistics is correct in assuming that the
performance-competence gap is substantial, then we should not expect
language models trained on performance data to acquire human-like
competence and reveal the nature of human competence. Importantly, the
antecedent assumption is highly controversial; many linguistic theories
do not postulate the existence of an insurmountable discrepancy between
the surface structure of language and the structures that subserve
language acquisition and use
\citep[e.g.,][]{culicoverSimplerSyntax2005, pinkerFacultyLanguageWhat2005, tomaselloConstructingLanguage2009, christiansenCreatingLanguageIntegrating2016}.
While proponents of these theories may not deny that some performance
constraints on behaviour can be dissociated from linguistic competence,
they do not assume that the latter is a radically different and minimal
set of rule systems that cannot be induced from performance data. If
that is the case, then studying language models under the lens of
behavioural tasks, diagnostic probes, and causal interventions might
give insights not only into human performance, but also into human
competence
\citep{linzenWhatCanLinguistics2019a, linzenSyntacticStructureDeep2021a}.

The principled stance against the relevance of language models to
theoretical linguistics can also be turned on its head. The sharp
competence/performance distinction postulated by mainstream generative
grammar is justified, at least in part, by negative claims about the
learnability of language from mere exposure to data. As we shall see in
Section~\ref{sec-in-principle-claims}, language models may challenge
those claims, by providing a potential existence proof for the success
of statistical learning without innate grammatical constraints
\citep{contreraskallensLargeLanguageModels2023, piantadosiModernLanguageModels2023}.
In turn, this might weaken the motivation for an absolute
performance/competence gap, and correspondingly increase the relevance
of language models to linguistic theory.

There is a further question about whether it makes sense to apply the
performance/competence distinction to language models themselves, and if
so, how the distinction manifests
\citep{firestonePerformanceVsCompetence2020, katzirWhyLargeLanguage2023}.
For humans, we can attribute many performance failures to temporary
recoverable factors and experimentally control these variables to reveal
competence. While model performance may be underestimated by inadequate
decoding methods \citep{huPromptbasedMethodsMay2023} and mismatched
experimental conditions
\citep{lampinenCanLanguageModels2023, cowleyFrameworkRigorousEvaluation2022, frankLargeLanguageModels2023},
this is not directly equivalent to the competence/performance
distinction, at least as the distinction is framed by generative
linguists. A closer analogue might be found in (a) cases of
misrepresentation of syntactic features revealed by causal interventions
on probed activations
\citep{giulianelliHoodUsingDiagnostic2018, hardingOperationalisingRepresentationNatural2023},
and (b) cases in which multiple circuits are competing for influence on
model behaviour
\citep{zhongClockPizzaTwo2023, milliereAnthropocentricBiasPossibility2024}.
Such cases may provide evidence that a model has the capacity to
represent a given feature or perform a given computation, even though
this capacity is not \emph{always} reflected in its observable
behaviour.

If there is a meaningful distinction between performance and competence
when it comes to language models, we might wonder whether it is always
appropriate to discount performance errors as irrelevant to the
assessment of competence in the case of humans, but not in the case of
models. This relates to the comparative bias that
\citet{bucknerMorganCanonMeet2013} termed ``anthropofabulation'': the
tendency to assess nonhuman performance against an inflated conception
of human competence \citep[see also][ and
\citet{milliereAnthropocentricBiasPossibility2024}]{bucknerBlackBoxesUnflattering2021}.
In the context of the evaluation of language models on syntactic tasks,
anthropofabulation might manifest itself through the expectation that
neural networks should achieve perfect or near-perfect performance
accuracy to be ascribed human-like competence in the domain, while
performance mistakes are ignored in the evaluation of human competence.
If there are contingent limitations on the externalization of syntactic
competence in language models (e.g., interference from competing
circuits in some circumstances), then it might be reasonable to downplay
some of their performance errors in human-machine comparisons.
Alternatively, if the gap between human performance and competence is
narrower than typically assumed by generative linguistics, then human
performance errors in appropriately matched experimental conditions
ought to be taken into account.

\subsection{In-principle claims about competence and
learnability}\label{sec-in-principle-claims}

A traditional charge against connectionist models is that they have
fundamental and insurmountable limitations that make them inadequate as
models of cognition or linguistic competence, unless they merely
implement classical symbolic structures
\citep{fodorConnectionismCognitiveArchitecture1988, pinkerLanguageConnectionismAnalysis1988, marcusRethinkingEliminativeConnectionism1998}.
Generative linguists, in particular, hold that statistical and
usage-based approaches to language modelling face systematic
limitations, because their reliance on linear string order cannot
account for the hierarchical structure dependence of syntactic rules
\citep{everaertStructuresNotStrings2015}.

These theoretical claims have inspired research on in-principle
capabilities and limitations of language models architectures. On the
one hand, \citet{hahnTheoreticalLimitationsSelfAttention2020} found that
the self-attention mechanism in Transformers cannot model periodic
finite-state languages or hierarchical structure unless the number of
layers or heads increases with input length.
\citet{deletangNeuralNetworksChomsky2022} also tested various neural
networks architectures on a battery of sequence prediction tasks
designed to span the Chomsky hierarchy of formal grammars
\citep{chomskyCertainFormalProperties1959}. They found that contemporary
language models architectures like LSTMs and Transformers do not neatly
fit into the hierarchy. LSTMs can solve some simple context-sensitive
tasks but fail on most, while Transformers fail on many regular tasks.
On the other hand, \citet{yaoSelfAttentionNetworksCan2021} showed that
Transformers can process bounded hierarchical formal languages that
adequately capture the bounded hierarchical structure of natural
language better than their unbounded counterparts, and that they have a
memory advantage over RNNs despite lacking a built-in recursive
mechanism. Related work found that Transformers can effectively learn
``shortcut'' solutions that replicate the computations of recurrent
models in a single pass \citep{liuTransformersLearnShortcuts2022}, and
that they have an inherent simplicity bias that shapes their
generalization capabilities beyond what classical theory predicts,
allowing success on tasks they should theoretically fail at
\citep{bhattamishraSimplicityBiasTransformers2023}.

In practice, there is converging empirical evidence that
Transformer-based language models are capable of processing bounded
hierarchical phrase structure and recursion in a naturalistic context
\citep{muellerColoringBlankSlate2022, lampinenCanLanguageModels2023, allen-zhuPhysicsLanguageModels2023, begusLargeLinguisticModels2023, dabkowskiLargeLanguageModels2023, zhaoTransformersParsePredicting2023}.
As discussed in Section~\ref{sec-syntax}, a wealth of experiments show
that language models are sensitive to, and encode information about,
many morphological and syntactic features -- including number agreement,
constituency, long-distance dependencies, coreference and anaphora,
among others. Together with their resounding success on traditional NLP
tasks and fluent natural language generation, these results have
prompted a reappraisal of learnability claims in theoretical
linguistics.

\citet{goldLanguageIdentificationLimit1967} was highly influential in
framing the problem of language acquisition as one of grammatical
inference. Gold's theorem formally shows that for many common classes of
languages, no learner can be guaranteed to eventually converge on the
correct grammar for the target language based only on positive example
sentences. Importantly, this theorem only shows limitations of a
specific formal model of learning based solely on positive example
sentences; it does not directly model real-world language acquisition,
where its strong assumptions are very unlikely to hold. Nonetheless, it
has been widely (mis)interpreted as demonstrating that language
acquisition is only possible if the learner's hypothesis space is
heavily constrained by innate knowledge
\citep{clarkLinguisticNativismPoverty2010}.

This strong nativist claim about the in-principle learnability of
grammar on the basis of mere exposure to data has endured in generative
linguistics
\citep[e.g.,][17-20]{carnieSyntaxGenerativeIntroduction2021}.\footnote{Somewhat
  confusingly, the defence of this claim on the basis of formal
  learnability results such as Gold's theorem is sometimes referred to
  as the ``poverty of the stimulus'' (PoS) argument, although the
  canonical -- and less implausible -- version of the PoS argument, as
  we shall see, rests on empirical evidence from developmental
  psycholinguistics \citep[see][ for a thorough
  discussion]{pearlPovertyStimulusTears2022}.} Aside from being
unwarranted by formal learnability results under plausible learning
assumptions, this strong negative claim has come under pressure from
positive evidence of effective statistical learning. In this context,
neural network models can provide an existence proof of learnability
that undermines in-principle claims; as
\citet{elmanRethinkingInnatenessConnectionist1996} put it,
``connectionist simulations of language learning can be viewed as
empirical tests of learnability claims'' (p.~385). While simple
recurrent neural networks of the 1990s did not quite live up to that
promise, the successes of modern language models have been put forward
as conclusive evidence against strong learnability claims \citep[see
also][ for a review of similar claims from computational
linguists]{baroniProperRoleLinguistically2022}:

\begin{quote}
``The rise and success of large language models undermines virtually
every strong claim for the innateness of language that has been proposed
by generative linguistics.''
\citep[1]{piantadosiModernLanguageModels2023}
\end{quote}

\begin{quote}
``{[}Language{]} models provide an existence proof that the ability to
produce grammatical language can be learned from exposure alone without
language-specific computations or representations.''
\citep[6]{contreraskallensLargeLanguageModels2023}
\end{quote}

\begin{quote}
``The best present-day LLMs {[}large language models{]} clearly have
substantial competence\ldots{} {[}They{]} induce the causal structure of
language from purely distributional training.''
\citep[19]{pottsCharacterizingEnglishPreposing2023}.
\end{quote}

Insofar as they can learn structure from strings, modern language models
do plausibly undermine strong in-principle learnability claims. This
does not entail, however, that they actually learn language like humans
do, or even that they could do so in a learning environment comparable
to those children are immersed in. For language models to constrain
hypotheses about human language acquisition and challenge linguistic
nativism beyond strong learnability claims, we need additional evidence
from experiments that carefully control learning parameters based on
developmental considerations.

\subsection{Language models as model learners}\label{sec-model-learners}

There are two major criticisms of the viability of language models as
models of language acquisition. The first has to do with the hypothesis
that as statistical models with weak inductive biases, language models
can learn and process both natural and ``impossible'' languages with
equal proficiency
\citep{moroLargeLanguagesImpossible2023, chomskyNoamChomskyFalse2023}.
Impossible languages, in this context, refer to linguistic structures
that allegedly violate fundamental principles of human language, such as
hierarchical organization and recursion -- properties believed to be
innate to the human language faculty by generative linguists
\citep{moroImpossibleLanguages2016}.
\citet{mitchellPriorlessRecurrentNetworks2020} showed that RNNs could
successfully learn and perform well on subject-verb agreement tasks in
impossible languages, including those with reversed word order, repeated
tokens, and even randomly shuffled sentences. These findings suggest
that neural language models can indeed acquire and process linguistic
structures that are considered impossible for human learners. This
ability to handle both possible and impossible linguistic structures
without distinction is seen by generative linguists as evidence that
language models lack the intrinsic constraints that shape human
linguistic competence and guide natural language acquisition.

However, recent work by \citet{kalliniMissionImpossibleLanguage2024}
challenges this criticism. Their experiments with GPT-2 models on a
spectrum of impossible languages showed that these models do not learn
impossible languages as efficiently as natural ones. They found clear
distinctions in model perplexities, with models trained on
nondeterministic sentence shuffles performing worst, followed by
deterministic shuffles and local shuffles. By contrast, models trained
on unshuffled English consistently achieved the lowest perplexities.
Furthermore, their analysis showed that GPT-2 models preferred natural
grammar rules and developed human-like solutions even for non-human
patterns. This suggest that Transformer-based language models may in
fact have inductive biases that favour natural language structures, and
that we should not rule out their relevance to debates on language
acquisition on the grounds that they learn impossible languages with
equal facility.

The other major criticism of the relevance of language models to
theories of language acquisition is the discrepancy that typically
exists between their learning conditions and those of actual children.
Most strikingly, state-of-the-art language models like GPT-3 and GPT-4
learn from an inordinate amount of data -- hundred of billions to
trillions of words, representing a gap of four to five orders of
magnitude with the estimated language input of human children
\citep{frankBridgingDataGap2023, wilcoxBiggerNotAlways2024}. This
seriously undermines the relevance of large language models to debates
about human language acquisition.

One of the main empirical arguments in favour of linguistic nativism is
the so-called ``poverty of the stimulus'' (PoS) argument
\citep{chomskyAspectsTheorySyntax1965, berwickPovertyStimulusRevisited2011, pearlPovertyStimulusTears2022}.
At the heart of PoS is an induction problem: linguistic input data
available to children seem insufficient, on their own, to allow them to
acquire the correct linguistic generalization about the underlying
structures within a large hypothesis space. Yet developmental evidence
suggests that children make constrained generalizations to the correct
hypotheses quickly and uniformly across languages. This has led
generative linguists to conclude that children must have some innate
knowledge that allows them to bridge the gap between their limited input
data and linguistic generalizations.

The extent to which children's linguistic stimulus is as impoverished as
PoS assumes is debated
\citep{pullumEmpiricalAssessmentStimulus2002, clarkLinguisticNativismPoverty2010, chaterEmpiricismLanguageLearnability2015}.
There is also some notable variance in the quantity and quality of input
received by children in different cultural and socioeconomic
environments
\citep{huttenlocherLanguageInputChild2002, huangExploringSocioeconomicDifferences2017, bergelsonWhatNorthAmerican2019, cristiaChildDirectedSpeechInfrequent2019}.
Estimates of infant speech exposure range from as little as one hour to
as much as 3,300 hours of speech per year, which reflects uncertainties
about what constitutes meaningful input, cultural variations in
child-directed speech, and the impact of factors like background noise
\citep{coffeyDifficultyImportanceEstimating2024}. Nonetheless, one thing
is certain: in order to challenge the claim that innate knowledge is
required to solve the induction problem of language learning, artificial
model learners need to be trained on a realistic amount of data.

It should be noted that the input data and learning process of large
language models differ from those of children in several ways beyond
mere data quantity. Firstly, while children learn primarily through
speech in interactive social contexts, language models typically learn
from static text corpora.\footnote{It should be noted that speech
  language models are increasingly prevalent, suggesting that deep
  neural networks can effectively learn syntax from speech rather than
  tokenized written text
  \citep{lakhotiaGenerativeSpokenLanguage2021, begusBasicSyntaxSpeech2024}.}
In particular, children receive immediate feedback and corrections
during their language learning process, allowing them to adjust their
understanding based on communicative goals. Secondly, children learn
within a rich multimodal environment, where language is grounded in
sensorimotor experiences, perception, and social interaction; most
language models, by contrast, learn from text cues alone. Thirdly, the
content and structure of the linguistic data differ significantly.
Child-directed speech, often simplified and contextualized, contrasts
sharply with the diverse and complex datasets language models are
trained on, which include sources such as Wikipedia, books, web pages.
In addition, state-of-the-art language models are trained not just on
natural language, but also on substantial amounts of computer code,
which has no equivalent in a child's linguistic experience.

Carefully controlling these variables to select and train better model
learners could in principle constrain hypotheses regarding the necessary
and sufficient conditions for language learning in humans
\citep{warstadtWhatArtificialNeural2022, pearlmodelingSyntacticAcquisition2023, connellWhatCanLanguage2024}.
Results obtained from models whose learning scenarios more closely match
hypotheses about human learning are more likely to generalize to real
human learners. That said, not all discrepancies in learning conditions
are problematic, particularly in cases where artificial learners are at
a disadvantage compared to humans (e.g., by lacking access to multimodal
input). If models can still learn the target linguistic knowledge in
spite of the disadvantage, then \emph{a fortiori} humans should be able
to learn it without the disadvantage. It is far harder to establish
negative results: just because a language model fails to learn some
target knowledge does not mean that humans cannot either.

A growing number of studies set out to test language models in more
human-like conditions, mainly by limiting the amount of data they are
trained on to a developmentally plausible quantity, and making their
content more similar to child-directed speech.
Table~\ref{tbl-learnability} summarizes the main findings from recent
studies investigating the learnability of various linguistic phenomena
in language models trained on developmentally plausible data.

\begin{longtable}[]{@{}
  >{\raggedright\arraybackslash}p{(\columnwidth - 4\tabcolsep) * \real{0.6159}}
  >{\raggedright\arraybackslash}p{(\columnwidth - 4\tabcolsep) * \real{0.1258}}
  >{\raggedright\arraybackslash}p{(\columnwidth - 4\tabcolsep) * \real{0.2583}}@{}}
\toprule\noalign{}
\begin{minipage}[b]{\linewidth}\raggedright
\textbf{Linguistic phenomena tested}
\end{minipage} & \begin{minipage}[b]{\linewidth}\raggedright
\textbf{Implication}
\end{minipage} & \begin{minipage}[b]{\linewidth}\raggedright
\textbf{Reference}
\end{minipage} \\
\midrule\noalign{}
\endfirsthead
\toprule\noalign{}
\begin{minipage}[b]{\linewidth}\raggedright
\textbf{Linguistic phenomena tested}
\end{minipage} & \begin{minipage}[b]{\linewidth}\raggedright
\textbf{Implication}
\end{minipage} & \begin{minipage}[b]{\linewidth}\raggedright
\textbf{Reference}
\end{minipage} \\
\midrule\noalign{}
\endhead
\bottomrule\noalign{}
\endlastfoot
Various syntactic and semantic features & Moderately positive &
\citet{zhangWhenYouNeed2020} \\
Syntactic categories, semantic categories, determiner-noun agreement,
verb argument structure & Moderately positive &
\citet{wangFindingStructureOne2023} \\
Filler-gap dependencies and island constraints & Positive &
\citet{wilcoxUsingComputationalModels2023} \\
General linguistic knowledge & Positive &
\citet{samuelTrained100Million2023} \\
Generation of coherent and grammatically correct text & Positive &
\citet{eldanTinyStoriesHowSmall2023} \\
Question formation and passivisation & Moderately positive &
\citet{muellerHowPlantTrees2023} \\
Subject-verb agreement, wh-questions, relative clauses & Moderately
positive & \citet{evansonLanguageAcquisitionChildren2023} \\
Formation of yes/no questions & Negative &
\citet{yedetoreHowPoorStimulus2023} \\
Article+Adjective+Numeral+Noun construction & Positive &
\citet{misraLanguageModelsLearn2024} \\
Exceptions to passivisation & Moderately positive &
\citet{leongTestingLearningHypotheses2024} \\
\caption{Overview of studies investigating the learnability of various
linguistic phenomena in small language models trained on a
developmentally plausible amount of data. The ``Implication'' column
indicates whether each paper's findings generally support the
learnability of the corresponding
phenomenon.}\label{tbl-learnability}\tabularnewline
\end{longtable}

\citet{zhangWhenYouNeed2020} investigated how much pretraining data is
needed for language models to acquire linguistic knowledge by probing
RoBERTa models trained on varying amounts of data (1M to 30B words).
Using multiple probing methods including classifier probing, minimum
description length probing, and unsupervised grammaticality judgments on
BLiMP, they found that most syntactic and semantic features can be
learned with only 10-100M words of pretraining data. However,
performance on downstream natural language understanding tasks continued
to improve with billions of words, suggesting that skills beyond basic
linguistic knowledge are required for these tasks and take much more
data to acquire.

In the same vein, \citet{samuelTrained100Million2023} trained BERT-like
models on the 100-million-word British National Corpus and evaluated
them using a variety of linguistic probing tasks, downstream NLP
benchmarks, and BLiMP. Their models outperformed the original BERT
trained on 3.3 billion words, suggesting that carefully curated smaller
datasets can be more effective than larger web-crawled corpora. They
also found that models trained on the British National Corpus
substantially outperformed those trained on a random 100-million-word
subset of Wikipedia and books, confirming the importance of data quality
over quantity.

\citet{wangFindingStructureOne2023} trained models on a subset of
linguistic input from a single child's first two years, using
transcripts of child-directed speech from a head-mounted camera. Even
with this limited data, networks learned to differentiate syntactic
categories (e.g., nouns vs.~verbs) and semantic categories (e.g.,
animals vs.~clothing), and showed some sensitivity to linguistic
phenomena like determiner-noun agreement. However, the networks
struggled with more complex phenomena requiring longer-distance
dependencies, such as subject-verb agreement. Adding visual information
provided only incremental improvements in word prediction, especially
concrete nouns, without fundamentally altering the linguistic
representations. In follow-up work,
\citet{qinSystematicInvestigationLearnability2024a} trained six
different neural network architectures (including LSTMs and
Transformers) on five datasets, including three single-child linguistic
input corpora and two baseline corpora. They evaluated the models using
linguistic acceptability tests, visualizations of word embeddings, and
cloze tests, finding that models trained on single-child datasets
consistently learned to distinguish syntactic and semantic categories
and showed sensitivity to certain linguistic phenomena, performing
similarly to models trained on larger aggregated datasets.

\citet{muellerHowPlantTrees2023} assessed how pre-training data and
model architecture affect the emergence of syntactic inductive biases in
Transformer language models. They fine-tuned models on syntactic
transformation tasks (question formation and passivisation) and
evaluated out-of-distribution generalization to test for hierarchical
vs.~linear rule learning. They found that model depth was more important
than width or total parameter count for acquiring hierarchical biases.
Additionally, they showed that pre-training on simpler language like
child-directed speech (5M words) induced stronger syntactic biases than
pre-training on much larger amounts (up to 1B words) of more complex
text like Wikipedia or web crawl data.

\citet{wilcoxUsingComputationalModels2023} tested whether language
models -- including RNNs and Transformers -- could learn English
filler-gap dependencies and island constraints. They found that models
trained on corpora as small as 90 million words could acquire not only
basic filler-gap dependencies, but also their hierarchical restrictions,
unboundedness, and most island constraints.

\citet{eldanTinyStoriesHowSmall2023} generated a dataset of short
children stories containing only words that a typical three to four
year-old child would understand. After training very small language
models (under 10 million parameters) on this dataset, they found that
these models could produce fluent, consistent, and diverse stories with
good grammar and some reasoning ability.

Looking at learning dynamics across model learners and humans can also
yield some insights. \citet{evansonLanguageAcquisitionChildren2023}
trained 48 GPT-2 models from scratch on a small dataset of Wikipedia
articles and evaluated their linguistic abilities using 96 probes from
established benchmarks at regular intervals during training. They found
that the models learned linguistic skills in a consistent order across
random seeds, with learning occurring in parallel but at different rates
for different skills. Comparing a subset of syntactic probes to data
from 54 children aged 18 months to 6 years, they observed that the order
of acquisition for simple sentences, wh-questions, and relative clauses
was the same in the models as in children, though the models relied more
on heuristics than true syntactic understanding for the most complex
structures.

Ongoing efforts such as the BabyLM Challenge
\citep{warstadtCallPapersBabyLM2023, warstadtInsightsFirstBabyLM2024}
are bringing more evidence to bear on whether language models trained on
child-directed speech can learn syntax as efficiently as actual human
children without built-in syntactic knowledge. The BabyLM Challenge was
explicitly designed to incentivize research on sample-efficient language
model pretraining using developmentally plausible data. Participants
were tasked with training language models on a corpus constructed to
mimic the linguistic input available to a child by early adolescence.
The first challenge included three tracks: \emph{Strict} and
\emph{Strict-Small}, which required using only the provided dataset of
100M and 10M words respectively, and \emph{Loose}, which allowed
additional non-linguistic data. The training corpus was carefully
curated to include child-directed speech, transcribed dialogues,
children's books, and other age-appropriate texts. Models were evaluated
on a range of tasks including syntactic judgment (BLiMP), natural
language understanding (GLUE), and generalization ability (MSGS).

The top-performing submissions to first BabyLM challenge achieved
results comparable to much larger language models on certain tasks
\citep{wilcoxBiggerNotAlways2024}. For instance, the best-performing
model in the \emph{Strict} track, achieved scores on the BLiMP syntactic
judgment task that were only about 3\% below human-level performance and
comparable to models trained on orders of magnitude more data
\citep{georgesgabrielcharpentierNotAllLayers2023}. In terms of strategy,
architectural tweaks proved to be more impactful, with models based on
the LTG-BERT architecture from \citet{samuelTrained100Million2023}
performing particularly well. Interestingly, curriculum learning --
learning in a specific order -- generally showed only marginal
improvements over baselines, challenging common assumptions about the
benefits of structured learning for language models in limited data
scenarios.

Beyond training language models on more developmentally plausible data,
\citet{warstadtWhatArtificialNeural2022} suggest that specific features
of the model's learning conditions could be purposefully removed (or
``ablated'') to test whether they are really needed for learning. For
example, one could remove all triply nested sentences from the training
data, to test if this input is necessary for the model to acquire
knowledge of subject-verb number agreement in deeply embedded clauses.
If the model succeeds in acquiring the target knowledge despite lacking
the ablated advantage, it provides an existence proof that the knowledge
is learnable without it. By manipulating model assumptions and training
data, we can test which conditions are actually required for human-like
learning of linguistic rules and generalizations. We can also analyse
the internal representations that evolve to support model behaviour in
plausible learning scenarios through causal interventions, to constrain
hypotheses about representations subserving linguistic competence in
humans.

\citet{leongTestingLearningHypotheses2024} provide a good illustration
of this strategy. Using targeted interventions on the training data of
small language models, they test specific hypotheses about the sources
of evidence learners might use to learn exceptions to passivisation
(e.g.~``The meeting lasted one hour'' vs ``*One hour was lasted by the
meeting''). By manipulating factors like the frequency of verbs in
passive constructions or the semantic contexts in which verbs appear,
they were able to isolate and test the causal role of different types of
indirect evidence in the learning process. This methodology addresses a
major limitation of naturalistic studies of language acquisition, where
it is extremely difficult to control or measure a child's exact
linguistic input. Their findings suggest that the relative frequency of
verbs in active versus passive constructions plays a significant role in
how language models learn passive exceptions, while manipulations of
lexical semantics had less consistent effects. Importantly, they found
that frequency alone could not fully account for the models' judgments,
indicating that other sources of evidence are likely involved. Their
findings also show how language models can learn to make graded
acceptability judgments that correlate well with human intuitions, even
for subtle linguistic phenomena.

Similarly, \citet{patilFilteredCorpusTraining2024} trained language
models on corpora with specific linguistic constructions filtered out to
test generalization from indirect evidence. They applied this method to
both LSTM and Transformer models across a wide range of linguistic
phenomena evaluated by the BLiMP benchmark. Their results showed that
while Transformers achieved lower perplexity, both model types performed
equally well on linguistic generalization measures, with relatively
small impacts from filtering in most cases. This provide further
evidence that language models are capable of forming sophisticated
linguistic generalizations even without direct exposure to certain
constructions during training.

\citet{misraLanguageModelsLearn2024} used a similar strategy to
investigate whether language models trained on a 100-million-word corpus
could learn the rare Article-Adjective-Numeral-Noun (AANN) construction
in English. They systematically manipulated the training corpus by
removing AANNs and related constructions, then evaluated models on novel
AANN instances. They found that models could generalize to unseen AANNs
even without exposure to any during training, likely by leveraging
related frequent constructions. Additionally, they showed that models
exposed to more diverse AANN instances during training showed better
generalization, highlighting the importance of input variability in
learning rare phenomena.

While these results mostly point to challenges for nativist views, some
experiments with model learners in plausible learning scenarios found
less positive results. For example, \citet{yedetoreHowPoorStimulus2023}
trained LSTM and Transformer models on child-directed speech from the
CHILDES corpus (9.6 million words) to test whether they could learn the
hierarchical structure of English yes/no questions. They evaluated the
models using forced-choice acceptability judgments and a question
formation task, finding that both model types failed to acquire the
correct hierarchical rule. Instead, the models tended to generalize
based on linear order or lexical specificity, even when pre-trained on
next-word prediction. These results suggest that stronger constraints
might be needed to induce hierarchical syntax from the realistic input
children receive.

To shed further light on this question,
\citet{mccoymodelingRapidLanguage2023} used a technique called inductive
bias distillation to endow a neural network with the strong inductive
biases of a Bayesian model. The resulting network exhibited data
efficiency comparable to the Bayesian model in learning new formal
languages from few examples, but was also able to effectively learn
aspects of English syntax from a naturalistic corpus of child-directed
speech. Notably, it outperformed standard neural networks on targeted
evaluations of syntactic phenomena like dependencies, agreement, and
reflexives. Thus, neural networks can learn meaningful generalizations
given suitable inductive biases, although the biases of vanilla
architectures might not be sufficient to model language acquisition
adequately.

This does not entail that good model learners should be endowed with the
kind of \emph{language-specific} inductive biases that generative
linguists deem necessary for language acquisition. In fact, it is not
clear that such strong biases would lead to better learning. An
interesting study by \citet{papadimitriouPretrainJustStructure2023}
investigated which structural biases allow Transformers to achieve
excellent performance on natural language modelling without explicit
syntactic supervision. By pretraining Transformers on artificial
languages exhibiting specific structures like recursion or
context-sensitivity before fine-tuning on English text, they were able
to manipulate the models' inductive biases in a controlled fashion. They
found that both recursive and non-recursive structural biases improve
English learning over a random baseline, with context-sensitivity
providing the best inductive bias over constituency recursion. These
results show that that Transformers can acquire languages beyond
finite-state regular grammars given appropriate inductive biases,
without restrictions to only context-free or context-sensitive
languages. While not directly confirming claims about human language
acquisition, the finding that non-recursive dependencies aid learning
better than recursion challenges theories positing recursion as the core
syntactic bias.

It is also worth emphasizing that modern deep neural network
architectures are not \emph{tabulae rasae}, but have distinct inductive
biases \citep{baroniProperRoleLinguistically2022}. Models with different
architectures trained on the same data may generalize (or fail to
generalize) in different ways. While vanilla LSTMs and Transformers lack
\emph{language-specific} innate knowledge, they have more domain-general
biases that go a long way towards explaining their success (and
limitations) on language modelling tasks. Their ability to learn
language in plausible learning scenarios may undermine linguistic
nativist accounts of PoS, but it does not necessarily undermine the idea
that nontrivial inductive biases are required for language to be
acquired. As such, it is quite natural for moderate nativists and
moderate empiricists to meet somewhere in the middle.\footnote{Many
  proponents of universal grammar agree that innate knowledge is not
  sufficient to explain language acquisition. Statistical learning is an
  important component of ``innately guided learning'', where universal
  grammar may constrain which statistical cues the learner should attend
  to
  \citep{yangUniversalGrammarStatistics2004, pearlHowStatisticalLearning2021, dupreEmpiricismSyntaxOntogeny2021}.
  One key disagreement, however, is whether the innate component of
  language learning (i.e., inductive biases) should be domain-general of
  language-specific
  \citep{clarkLearnability2015, chaterEmpiricismLanguageLearnability2015}.
  Insofar as vanilla language model architectures do not have
  language-specific inductive biases, their tentative success in
  realistic learning scenarios may count as evidence against a strongly
  modular language faculty as postulated by Chomsky.}

Overall, the work reviewed in this section lends plausibility to the
claim that language models trained in more realistic learning scenario
could in principle constrain theorizing about language acquisition, and
particularly PoS-style arguments for particular syntactic phenomena.
While there certainly remains significant differences between the
mechanisms and conditions in which language models learn compared to
human children, the cognitive plausibility of model learners should be
viewed as a graded concept, evaluated comparatively across specific
several dimensions, rather than as a binary property that models either
possess or lack \citep{beinbornCognitivePlausibilityNatural2024}. For
now, however, the claim that language models refute Chomsky's approach
to language \citep{piantadosiModernLanguageModels2023} remains somewhat
premature. Strong learnability claims do not hold up very well to
scrutiny, but evidence from model learners against weaker nativist
claims is still tentative.

\subsection{Language models as scientific
models}\label{sec-scientific-models}

Deep neural networks are increasingly treated as promising computational
models of human cognition in various domains, including vision science
\citep{cichyDeepNeuralNetworks2019, doerigNeuroconnectionistResearchProgramme2023}.
However, the status of neural networks as scientific models is
controversial. One common view is that predictive performance on
benchmarks is insufficient for neural networks to be scientifically
adequate explanations of a target cognitive phenomenon
\citep{wichmannAreDeepNeural2023}. Theoretical linguistics, particularly
in the generative tradition, tends to favour explanation over
prediction. On this view, linguistic explanation aims to provide deep,
abstract accounts of linguistic phenomena, often focusing on competence
rather than performance. Prediction, on the other hand, involves using
models to forecast linguistic behaviour or patterns, often based on
statistical approaches.

There are reasons to question this sharp dichotomy. As
\citet{egreExplanationLinguistics2015} emphasizes, prediction is equally
applicable in linguistics as in other empirical sciences, and any
non-trivial descriptive generalization in linguistics will be predictive
if testable on new cases.
\citet{nefdtPhilosophyTheoreticalLinguistics2024} goes further, arguing
that prediction is essential for scientific explanation in linguistics,
and that computational approaches focused on prediction can offer
valuable insights into theoretical questions. He suggests that the
neglect of prediction stems partly from historical reactions against
logical positivism and partly from the ``Galilean style'' in generative
linguistics that emphasizes abstract explanation over empirical
adequacy. The integration of predictive models, such as deep neural
networks, into theoretical linguistics is a way to bridge this divide.

On the opposite end of the spectrum, some have argued that language
models should not only be treated as \emph{bona fide} linguistic
theories, but as the best ones we have
\citep{piantadosiModernLanguageModels2023, ambridgeLargeLanguageModels2024}.
For example, \citet{baroniProperRoleLinguistically2022} argues that
language models can be viewed as algorithmic theories making linguistic
predictions. Specifically, he proposes to view an untrained language
model as equivalent to a theory defining a \emph{space of possible
grammars}; that space will look quite different depending on model
architecture (e.g., LSTMs vs.~Transformers) and parameter count. After
training on language-specific data, the model can be viewed as a
\emph{grammar} -- a system that can predict whether any sequence is
acceptable to an idealised speaker of the language. For this framing to
be viable, however, Baroni emphasises that the selection of model
architecture in experiments must be linguistically-motivated, and that a
greater mechanistic understanding of trained models is needed.
Ultimately, the field could move beyond testing language models on
well-known patterns such as subject-verb number agreement, to using them
to make prediction about previously unexplored patterns. In particular,
they seem apt to model fuzzy and probabilistic aspects of language
better than elegantly concise linguistic theories focused on algebraic
recursion.

One key issue with this proposal is whether the notorious opacity of
neural networks, including language models, should be seen as a
fundamental impediment to their ability to generate scientific
explanations. Explanatory models in cognitive science often take the
form of mathematical or computational models that encode theoretical
constructs and hypotheses about mechanisms
\citep{forstmannModelBasedCognitiveNeuroscience2015}. From this
perspective, the lack of simplicity, transparency, and theoretical
grounding of deep neural networks appears to undermine them as
explanatory models. In the linguistic domain, language models do not
compare favourably with respect to these particular criteria to the kind
of minimal and interpretable models provided by generative grammar.

However, this view rests on controversial assumptions in the philosophy
of science about desiderata for explanatory models. For example,
\citet{sullivanUnderstandingMachineLearning2022} argues that
implementational opacity need not be an impediment for neural networks
to provide understanding of real-world phenomena. One does not need to
fully understand the model itself in order to use it to understand its
target. Rather, it is uncertainty about whether models accurately
represent real systems -- called \emph{link uncertainty} -- that
prominently hinders understanding. Links between models and target
phenomena can be strengthened through rigorous scientific validation
providing empirical evidence that opaque model mechanisms reflect real
causal dependencies. One can see ongoing research on language models in
computational linguistics as progressing in that direction.

Whether reducing link uncertainty between opaque neural network models
and target phenomena is sufficient to provide genuine explanatory
understanding is debated. A more stringent requirement would include
some understanding of the model itself -- that is, reducing model
opacity in addition to link uncertainty
\citep{razImportanceUnderstandingDeep2022}. A common motivation for this
requirement is the suspicion that deep neural networks might rely on
spurious correlations even if they appear to capture genuine
dependencies in their explanatory target. However, research using causal
methods and mechanistic interpretability techniques
(Section~\ref{sec-interventional-studies}) is making headway in
understanding how language models learn and represent linguistic
features.

Another important aspect of scientific models is that they allow for
\emph{surrogative reasoning} about their explanatory targets; that is,
studying the model itself allows researchers to draw inferences about
the target system \citep{nguyenScientificRepresentation2022}.
Surrogative reasoning is what allows scientists to gain knowledge about
real-world systems by investigating simplified or idealized model
systems. The use of language models as model learners
(Section~\ref{sec-model-learners}) is an example of surrogative
reasoning; it allows computational linguists to draw inferences about
the relative importance of various inductive biases and properties of
the learning environment for human language acquisition.

These considerations provide tentative support against a merely
instrumentalist view of language models, on which they should only be as
viewed tools for prediction rather than models for explanation
\citep{katzirWhyLargeLanguage2023}. The case for viewing language models
as explanatory models is perhaps strongest for the study of language
acquisition with artificial learners, where the models' parameters and
environment are carefully controlled and informed by developmental
psycholinguistics. The case for language models as scientific models of
adult linguistic competence is perhaps more controversial, as it depends
on aforementioned assumptions about the performance-competence gap
\citep{dupreWhatCanDeep2021}. However, these two projects are not
orthogonal; if experiments with artificial model learners undermine
linguistic nativism, this might in turn weaken the case for a wide gap
between performance and competence, and correspondingly increase the
relevance of language models to arbitrate hypotheses about linguistic
competence.

At the very least, language models designed and trained with cognitive
and developmental plausibility in mind could in principle furnish
\emph{how-possibly} explanations of aspects of language acquisition or
linguistic competence. How-possibly explanations are possible
explanations of a phenomenon under certain plausibility constraints
\citep{bokulichHowTigerBush2014}. Scientific models can provide evidence
for how-possibly explanations by supporting judgments about the
possibility of explanatory relationships
\citep{verreault-julienHowCouldModels2019}. Importantly, this allows
highly idealized models to still contribute to how-possibly explanations
about real-world possibilities. The extent to which language models can
support such explanations of language acquisition or linguistic
competence arguably depends both on their cognitive plausibility and on
their interpretability. Ongoing efforts to develop more cognitively
plausible language models and interpret their computational structure
through causal interventions show promise in fulfilling that vision.
Unlike just-so stories, how-possibly explanations can provide a path to
scientific understanding through further investigation. For example,
some how-possibly explanations of the learnability of specific syntactic
features provided by language models in controlled learning scenarios
could in principle be put to the test in developmental psycholinguistics
-- or at least be evaluated against available cross-cultural evidence.
Conversely, nativist claims about the learnability of specific
constructions from sparse stimulus can be challenged by how-possibly
explanations derived from experiments with language models.

While principled research on language models might weaken or constrain
some PoS arguments, it's important to note that it doesn't necessarily
undermine all motivations for traditional linguistic theories.
Generative approaches like Minimalist syntax are not solely justified by
learnability considerations, but also by their ability to provide
elegant explanations for specific linguistic phenomena across languages.
For instance, the presence of expletive subjects in English sentences
like ``It's raining'' or ``There is a cat in the garden'' has been
accounted for through principles of case theory and the Extended
Projection Principle (EPP)
\citep{chomskyMinimalistProgram1995}.\footnote{I am grateful to Gabe
  Dupre for suggesting that example.} Case theory explains why
expletives are necessary in certain constructions to satisfy case
requirements, while the EPP stipulates that all clauses must have
subjects. While a language model trained on English text might correctly
produce such sentences, this alone doesn't explain why English requires
expletive subjects in these contexts in the same way that traditional
linguistic theories do. This highlights a crucial distinction between
prediction and explanation in linguistics. For language models to truly
challenge or replace traditional linguistic theories, they would need to
offer comparably insightful explanations for the cross-linguistic
patterns and constraints that motivate these theories. This presents an
important challenge and opportunity for researchers working on language
models: to develop methods for extracting explanatory principles from
these models that can account for linguistic phenomena in ways that
rival or surpass traditional theoretical approaches. Such efforts could
significantly enrich debates about the nature of linguistic explanation
and potentially bridge the gap between computational and theoretical
linguistics.

\section{Conclusion}\label{conclusion}

Artificial neural networks have come a long way since the much-maligned
connectionist models of yore. In the linguistic domain, modern language
models based on deep neural network architectures have achieved vastly
more success on virtually any natural language processing task than
symbolic models ever did. This progress calls for an honest reappraisal
of the relevance of artificial neural networks to linguistics. Given
their predictive learning objective, it is often assumed that language
models are limited to capturing human linguistic performance. There are,
however, good reasons to think they can be used to model key aspects of
language competence and acquisition. This requires careful experiments
where every variable -- from model choice to task design -- is informed
by linguistic theory.

Ongoing research on language models in computational linguistics has
frayed a path forward, providing a wealth of empirical evidence about
the linguistic abilities of neural networks, including those trained on
a realistic amount of linguistic input. These results suggest that
language models do acquire sophisticated linguistic knowledge and are
sensitive to hierarchical syntactic structure beyond surface heuristics.
Although this line of research has been largely ignored in mainstream
theoretical linguistics, it is increasingly plausible that it could
yield insights about linguistic competence and language acquisition that
could constrain hypotheses about the human case. This calls for a closer
collaboration between linguists and neural network researchers that does
not merely cater to engineering goals.\footnote{I am grateful to Emma
  Borg, Gabe Dupre, Chris Hill, Barry C. Smith, and Charles Rathkopf for
  comments on previous versions of this chapter.}

\bibliographystyle{plainnat}

\bibliography{bibliography}

\end{document}